\pdfoutput=1
\documentclass[10pt,twocolumn,letterpaper]{article}

\usepackage[pagenumbers]{cvpr} 

\definecolor{cvprblue}{rgb}{0.21,0.49,0.74}
\usepackage[pagebackref,breaklinks,colorlinks,allcolors=cvprblue]{hyperref}

\usepackage{amssymb}
\usepackage{amsmath}
\usepackage{graphicx}
\usepackage{booktabs}
\usepackage{multirow}
\usepackage{multicol}
\usepackage{colortbl}
\usepackage{array}        
\usepackage{amsfonts}
\usepackage{float}
\usepackage{bm}

\newtheorem{definition}{Definition}

\newcommand{\loss}{\mathcal{L}}

\def\paperID{2217} 
\def\confName{CVPR}
\def\confYear{2025}

\title{Proactive Gradient Conflict Mitigation in Multi-Task Learning: \\A Sparse Training Perspective}

\author{
\textbf{Zhi Zhang}\textsuperscript{1}, 
\textbf{Jiayi Shen}\textsuperscript{1}, 
\textbf{Congfeng Cao}\textsuperscript{1}, 
\textbf{Gaole Dai}\textsuperscript{2}, 
\textbf{Shiji Zhou}\textsuperscript{3}, 
\textbf{Qizhe Zhang}\textsuperscript{2}, \\
\textbf{Shanghang Zhang}\textsuperscript{2}\footnotemark[1], 
\textbf{Ekaterina Shutova}\textsuperscript{1}\footnotemark[1]\\
\textsuperscript{1}ILLC, University of Amsterdam, Netherlands \\
\textsuperscript{2}
State Key Laboratory of Multimedia Information Processing, \\School of Computer Science, Peking University, China\\
\textsuperscript{3}Department of Automation, Tsinghua University, China\\
{\tt\small zhangzhizz2626@gmail.com, j.shen@uva.nl, shanghang@pku.edu.cn, e.shutova@uva.nl}
}

\begin{document}
\maketitle

\renewcommand{\thefootnote}{\fnsymbol{footnote}}
\footnotetext[1]{Corresponding author.}

\begin{abstract}
Advancing towards generalist agents necessitates the concurrent processing of multiple tasks using a unified model, thereby underscoring the growing significance of simultaneous model training on multiple downstream tasks. A common issue in multi-task learning is the occurrence of gradient conflict, which leads to potential competition among different tasks during joint training. This competition often results in improvements in one task at the expense of deterioration in another.
Although several optimization methods have been developed to address this issue by manipulating task gradients for better task balancing, they cannot decrease the incidence of gradient conflict.
In this paper, we systematically investigate the occurrence of gradient conflict across different methods and propose a strategy to reduce such conflicts through sparse training (ST), wherein only a portion of the model's parameters are updated during training while keeping the rest unchanged. Our extensive experiments demonstrate that ST effectively mitigates conflicting gradients and leads to superior performance. Furthermore, ST can be easily integrated with gradient manipulation techniques, thus enhancing their effectiveness.



\end{abstract}

\section{Introduction}
\label{intro}
Attaining the status of a generalist agent necessitates addressing multiple tasks within a unified architecture, thereby emphasizing the significance of multi-task learning (MTL) \citep{zhang2021survey}, which involves concurrently acquiring proficiency in multiple tasks and striving for superior overall performance compared to learning these tasks separately. 

The primary concern for MTL lies in the phenomenon of task competition when the model is jointly trained by optimizing the average loss across all tasks. As a result, a subset of tasks demonstrates superior performance while others remain sub-optimized compared to their individual learning counterparts. 
One of the reasons behind it, from an optimization perspective, is gradient conflict (GC) \cite{PCGrad}, wherein the direction and magnitude of gradients between tasks differ significantly. This can result in the average gradient biasing towards optimizing one task while providing relatively smaller and sometimes even negative optimization for other tasks when updating the network \cite{PCGrad, liu2023famo}.

Numerous works have employed the gradient manipulation method to directly or indirectly adjust the gradients of tasks to mitigate the issue of gradient conflict in tasks. The former involves direct alteration of task gradients through manually designed criteria when conflicts arise \cite{PCGrad, Graddrop, CAGrad}, while the latter modifies task gradients by adjusting weights of loss for each task \cite{MGDA, imtlg, nash, liu2023famo}. Although these methods effectively modify the gradients conflicting with each other, they do not decrease the occurrence of conflicting gradients during training \cite{shi2023recon}. 

A simple approach to mitigate the occurrence of conflicting gradients is to convert those layers in which gradient conflict frequently arises into task-specific layers, thereby reducing the likelihood of gradient conflicts within the remaining shared layers \cite{shi2023recon}. 
However, this strategy introduces additional modules and disrupts the internal structure of the original model, resulting in increased computational costs. Furthermore, identifying frequently conflicting layers adds extra computational costs. This becomes prohibitively expensive as the model size continues to expand, and thus prompting our fundamental inquiry: 

\begin{figure*}[t]
\centering
\includegraphics[width=0.9\linewidth]{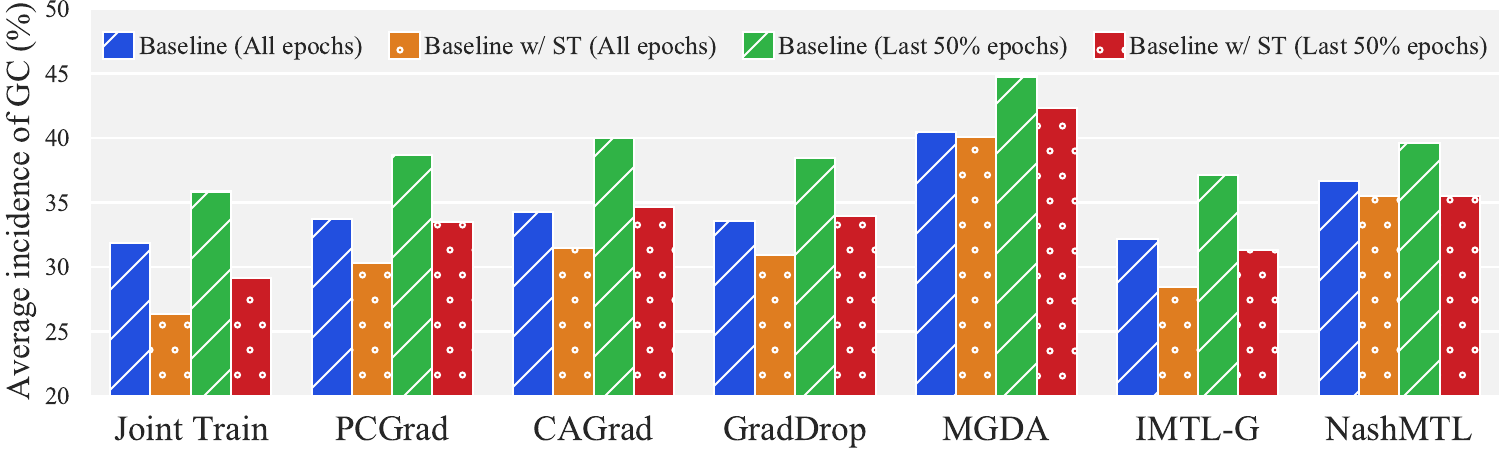}
\caption{
The average occurrence percentage of gradient conflict over epochs (all epochs/last 50\% epochs) during training on the SAM model with NYUv2 datasets is evaluated using various methods, including joint training and gradient manipulation techniques.
}
\label{fig:gc of different methods}
\end{figure*}

\vspace{+0.2cm}
(\textbf{\emph{Q}})~~\emph{Is there a universally applicable approach to proactively mitigate the occurrence of gradient conflicts as well as preserve architectural integrity for MTL?}
\vspace{+0.2cm}

To tackle this issue, we propose a novel perspective on mitigating gradient conflict in MTL, termed Sparse Training (ST), wherein a subset of parameters from the original model are selected to learn multiple tasks simultaneously while keeping the remaining parameters frozen. 
The intuition behind this lies in the reduction of a high-dimensional optimization problem to a low-dimensional one, which effectively alleviates the optimization complexity. Moreover, restricting the gradient updates of individual tasks to influence only a subset of parameters, rather than all parameters, effectively reduces potential interference between tasks.



Our key findings demonstrate that ST can effectively reduce the incidence of gradient conflict, particularly during the later stages of training, as illustrated in \cref{fig:gc of different methods}.  A summary of our contributions is as follows: \romannumeral1) We provide a novel perspective, sparse training, for proactively reducing the incidence of gradient conflict during training while keeping the architecture intact; \romannumeral2) Sparse training can be easily applied to improve various gradient manipulation methods by reducing the occurrence the gradient conflict over different datasets and architectures; \romannumeral3) In addition to conventional research that primarily focuses on smaller models (MTAN \cite{SegNet_attention} and SegNet \cite{badrinarayanan2017segnet}), we provide a comprehensive assessment of larger pre-trained models, including SAM \cite{chen2023sam}, ViT \cite{dosovitskiy2020image}, Swin Transformer \cite{liu2021swin}, using various gradient manipulation techniques, such as PCGrad \cite{PCGrad}, CAGrad \cite{CAGrad}, GradDrop \cite{Graddrop},  MGDA \cite{MGDA}, IMTL-G \cite{imtlg} and NashMTL \cite{nash}, to stimulate research in the field of sparse training for MTL. Our findings demonstrate that as the model size increases, the issue of gradient conflict becomes more exacerbated, as shown in \cref{fig: gc of different swin model size}, underscoring the significance of investigating the gradient conflict in large-scale models.

\section{Related work}
\label{sec: related work}
\paragraph{Multi-task optimization for MTL} 
The recent works~\cite{PCGrad, CAGrad, Graddrop, MGDA, imtlg, nash,liu2023famo} have achieved impressive results in addressing task imbalance issues in MTL by directly or indirectly modifying conflicting task gradients. Specifically, some works~\cite{PCGrad, CAGrad, Graddrop} propose to form a new update gradient at each training step by directly altering gradients based on certain criteria. 
Other works~\cite{MGDA, imtlg, nash,liu2023famo, kendall2018multi} learn dynamic loss scale to balance different tasks during training, and thus indirectly altering the gradient of tasks.
However, these methods only address GC when it occurs and do not proactively prevent it. In this paper, we sparsely train an MTL model, effectively reducing the incidence of GC.

\vspace{-0.3cm}
\paragraph{Training with subset of parameters}
Several methods have already been proposed in single-task learning.
Some of them select a subset of parameters based on a certain pre-defined rule, such as gradient \cite{zhang2023gradient, fu2023effectiveness} and magnitude of parameters \cite{lagunas2021block}. In addition to selecting parameters by hand design, the works in \cite{sanh2020movement, mostafa2019parameter, xu2021raise} automatically select the subset of parameters through optimization. Although sparse training has been extensively investigated in single-task learning, its application in MTL remains relatively unexplored. \citet{sun2020learning} and \citet{calandriello2014sparse} learn to share information between tasks using a sparse model instead of sparse training. Differently, we research the gradient conflict via the sparse training perspective.

\begin{figure*}[t]
\centering
\subfloat[Joint Train]{
    \includegraphics[width=0.21\linewidth, trim=3.8cm 3.1cm 1.2cm 2.5cm, clip]{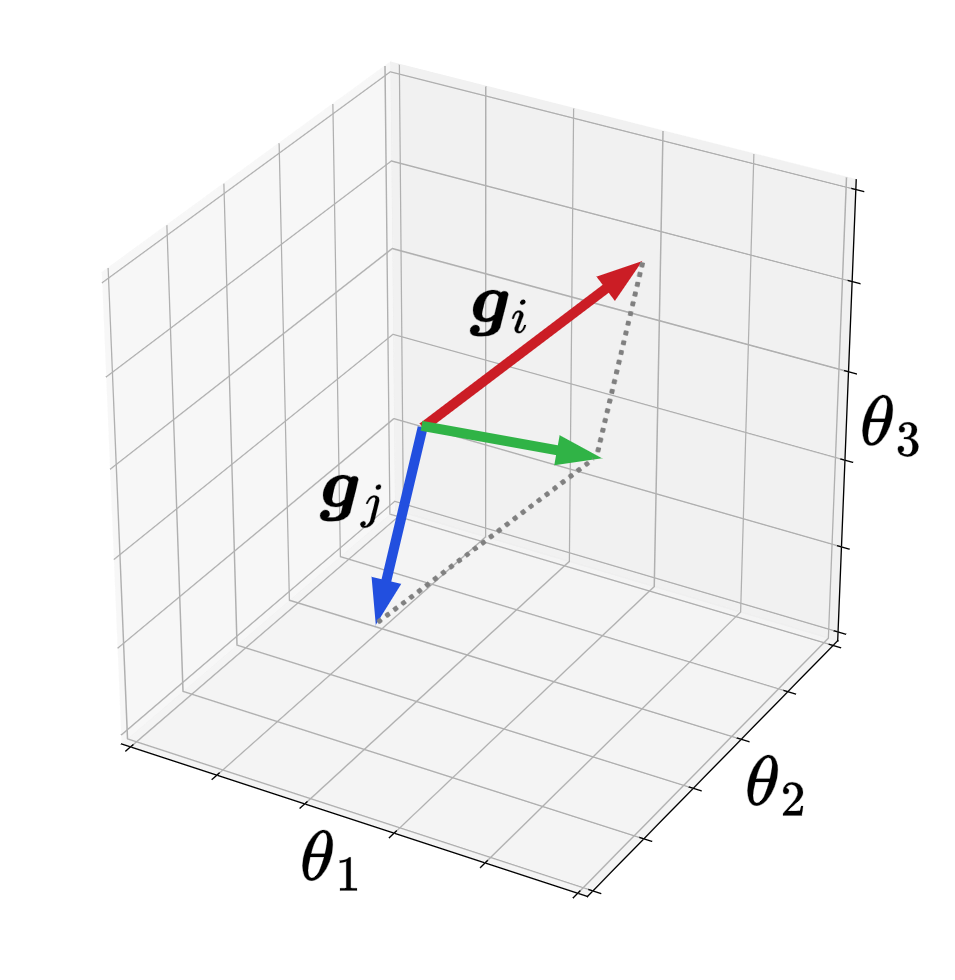}
    \label{fig:joint 3d}
}
\hfill
\centering
\subfloat[Joint Train w/ ST]{
    \includegraphics[width=0.21\linewidth, trim=3.8cm 3.1cm 1.2cm 2.5cm, clip]{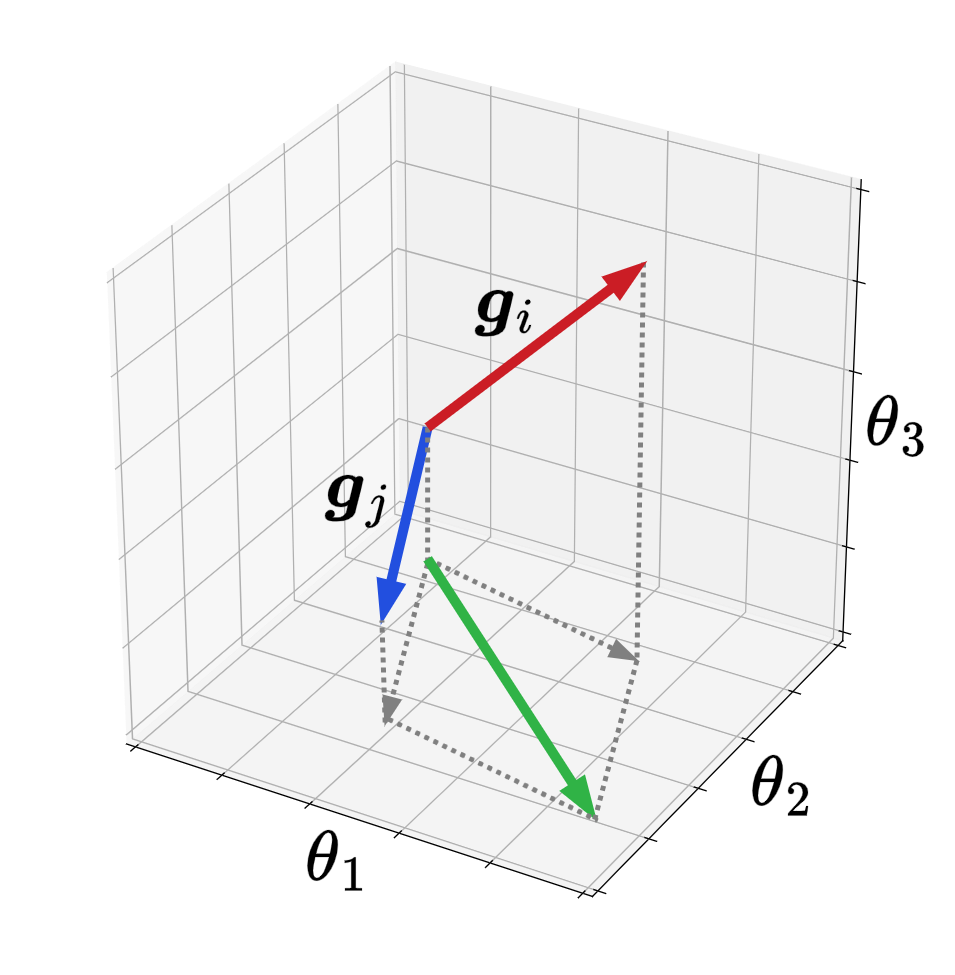}
    \label{fig:NSF 3d}
}
\hfill
\centering
\subfloat[PCGrad]{
    \includegraphics[width=0.21\linewidth, trim=3.8cm 3.1cm 1.2cm 2.5cm, clip]{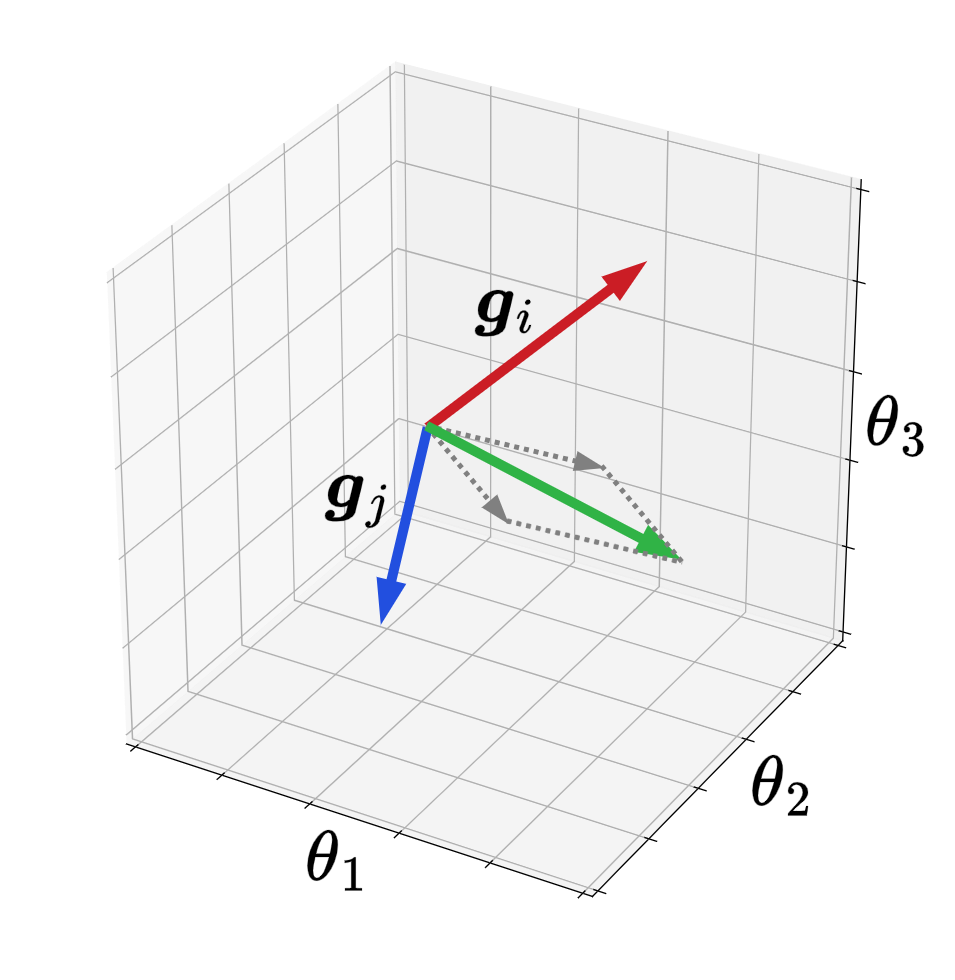}
    \label{fig:pcgrad 3d}
}
\hfill
\centering
\subfloat[PCGrad w/ ST]{
\includegraphics[width=0.21\linewidth, trim=3.8cm 3.1cm 1.2cm 2.5cm, clip]{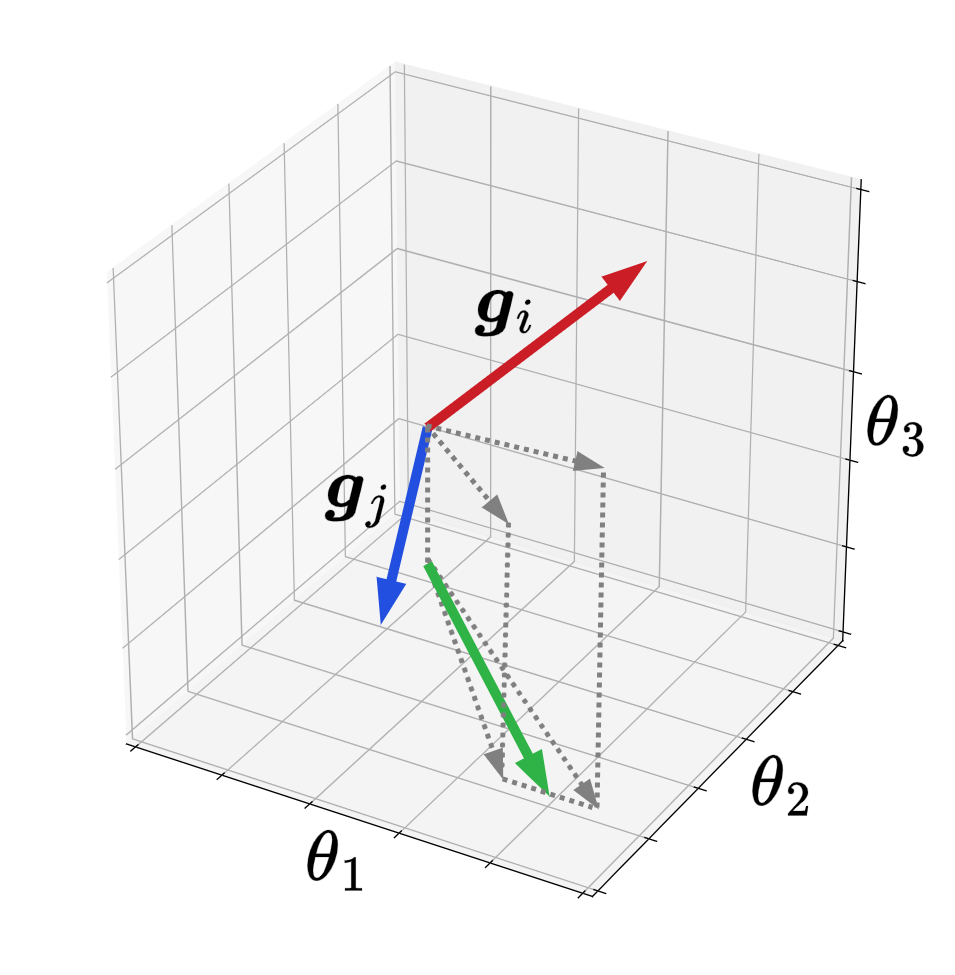}
    \label{fig:pcgrad nsf 3d}
}
\caption{Visualization of gradients change for different methods. $g_i$ and $g_j$ are two conflicting gradients, and the green arrow is the actual update vector. The process of sparse training can be interpreted as performing an orthographic/coordinate projection of conflicting gradients onto the subspace defined by the selected parameters, resulting in better alignment of the projected gradients.
}
\label{fig:3d gc}
\end{figure*}
\section{Approach}
\label{sec: approach}
\subsection{Background}
\paragraph{Multi-task learning (MTL)} aims to learn multiple tasks simultaneously within a single model. Formally, given  $\left\{\mathcal{T}_t\right\}_{t=1}^{T}$ tasks ($\geq 2$) and a model $\Theta$ with parameters $\Theta = (\theta_{\mathrm{sha}}, \theta_{\mathrm{sep}})$, where $\theta_{\mathrm{sha}}$ and $\theta_{\mathrm{sep}}$ are shared parameter with all tasks and task-specific parameters $\theta_{\mathrm{sep}}=\left\{\theta_{\mathrm{sep}}^t\right\}_{t=1}^{T}$ respectively, the commonly used optimization method for MTL (referred to as \textit{Joint Train}) is based on computing the average loss across all tasks with equal weights: 
\begin{equation}  
\label{eq:multi-objective}
    \Theta^{*} = \arg\min_\Theta \loss(\Theta),
\end{equation}
\begin{equation}
\loss(\Theta)=\loss(\theta_{\mathrm{sha}}, \theta_{\mathrm{sep}})=\frac{1}{T} \sum_{t=1}^{T} \loss_{t}(\theta_{\mathrm{sha}}, \theta_{\mathrm{sep}}^t)
\label{eq: mtl joint train loss}
\end{equation}
where each task $t$ is associated with a corresponding loss function $\loss_{t}(\theta_{\mathrm{sha}}, \theta_{\mathrm{sep}}^t)$. 
\vspace{-0.3cm}
\paragraph{Gradient conflict (GC)}
However, optimizing all tasks by aggregating their losses indiscriminately (\cref{eq: mtl joint train loss}) may lead to task competition, wherein certain tasks demonstrate improvement while others exhibit a decline compared to training them separately. From an optimization perspective, one of the reasons stems from conflicts in gradients. Formally, the update of task $\mathcal{T}_i$ may potentially exert a detrimental impact on another task $\mathcal{T}_j$, namely:
\begin{equation}
    \Delta\loss_j = \loss_j(\hat\theta_{\mathrm{sha}},\theta_{\mathrm{sep}}^{j}) - \loss_j(\theta_{\mathrm{sha}},\theta_{\mathrm{sep}}^{j}),
    \label{eq: gc}
\end{equation}
\begin{equation}
\hat\theta_{\mathrm{sha}} = \theta_{\mathrm{sha}}-\alpha \mathbf{g}_i 
\end{equation}
where $\mathbf{g}_i = \nabla_{{\theta}_{\mathrm{sha}}} \loss_i(\theta_{\mathrm{sha}}, \theta_{\mathrm{sep}}^{i})$ is the gradient of loss on task $\mathcal{T}_i$ with respect to $\theta_{\mathrm{sha}}$ and $\alpha$ is the learning rate. After the first-order Taylor approximation, \cref{eq: gc} can be expressed as $- \alpha \mathbf{g}_i \cdot \mathbf{g}_j + o(\alpha)$. Gradient conflict arises when $\mathbf{g}_i \cdot \mathbf{g}_j < 0$, leading to $\Delta\loss_j >0$, indicating that task $\mathcal{T}_i$ has a detrimental impact on task $\mathcal{T}_j$. Following \cite{PCGrad}, we provide the definition of gradient conflict:
\begin{definition}[Gradient Conflict]
\label{def:task_conflict}
If $\cos{\phi_{ij}}<0$, where $\phi_{ij}$ is the angle between gradients of two tasks $\mathbf{g}_i$ and $\mathbf{g}_j$ $(i\neq j)$, then $\mathbf{g}_i$ and $\mathbf{g}_j$ are deemed to exhibit gradient conflict.
\end{definition}
\vspace{-0.3cm}
\paragraph{Gradient manipulation} To alleviate the issue of gradient conflict, gradient manipulation methods adjust conflicting gradients based on specific criteria and utilize these modified gradients for model updating. Instead of updating the model on the average gradient in \cref{eq:multi-objective} and \cref{eq: mtl joint train loss}:
\begin{equation}
\nabla_{\theta_{\mathrm{sha}}}\loss(\Theta)=\frac{1}{T} \sum_{t=1}^{T}\nabla_{\theta_{\mathrm{sha}}}\loss_{t}(\theta_{\mathrm{sha}}, \theta_{\mathrm{sep}}^{t}),
\label{eq: gradient of joint train}
\end{equation}
the gradients of all tasks in gradient manipulation methods are modified as follows:
\begin{equation}
\nabla_{\theta_{\mathrm{sha}}}\loss_{\mathrm{gm}}(\Theta)=\frac{1}{T} \sum_{t=1}^{T}\boldsymbol{w}_t\nabla_{\theta_{\mathrm{sha}}}\loss_{t}(\theta_{\mathrm{sha}}, \theta_{\mathrm{sep}}^{t}),
\label{eq: gradient manipulation gradient}
\end{equation}
\begin{equation}
\boldsymbol{w}_t=f\left(\nabla_{\theta_{\mathrm{sha}}}\loss_{1}, \ldots, \nabla_{\theta_{\mathrm{sha}}}\loss_{T}\right)
\end{equation}
where $\boldsymbol{w}_{t}$ can be either pre-defined or dynamically computed for tasks via $f$ and thus achieve the aim of adjusting the task gradient \cite{liu2023famo, nash,MGDA, imtlg, PCGrad,Graddrop,CAGrad}. However, the results of our experiment suggest that these methods can only modify gradients when conflicts occur, rather than proactively reducing the occurrence of GC during training, compared with \textit{Joint Train}, as shown in \cref{fig:gc of different methods}.

\subsection{Sparse training for multi-task learning}
In this study, we investigate the gradient conflict commonly observed in multi-task learning from a novel perspective: sparse training, which selectively trains only a subset of the model parameters as opposed to full parameter training.
This perspective is based on the intuition that by converting a high-dimensional space optimization problem into a lower-dimensional one, the complexity of optimization can be effectively reduced.
Additionally, by limiting the impact of gradient updates to only a subset of parameters for each task instead of all parameters, potential interference between tasks can be mitigated.
\vspace{-0.3cm}
\paragraph{Sparse training (ST)} entails the initial parameter selection from the original model, and then updating only these parameters while keeping other parameters fixed during model training. 
To clarify potential misunderstandings regarding ST---often confused with sparse networks, where parameters are abandoned for model compression---we provide the following definition to ensure consistency and ease of understanding throughout this paper. 
\begin{definition}[Sparse Training]
\label{def:Sparse Training} 
Given a model $\Theta$ and a binary mask matrix $M$ indicating whether parameters in $\Theta$ are selected, where $M \in \mathbb{R}^{\lvert\Theta\rvert \times \lvert\Theta\rvert}$, $M_{ii} \in \{0,1\}$ and $M_{ij}=0$ $(\forall i\neq j)$, the model is updated by $\hat\Theta = \Theta - \alpha M \nabla_{\Theta}\loss(\Theta)$. We define this training strategy as sparse training.
\end{definition}
Typically, the model architecture in multi-task learning includes a shared encoder as a feature extractor with task-specific decoders for multiple tasks. Therefore, sparse training is used in the encoder, and full parameters training for the decoders. We detail how the mask is computed in section \cref{sec: neuron-based selection}. We now apply sparse training for multi-task learning (\textit{Joint Train}). The visualization of the gradient change can be viewed in \cref{fig:3d gc} and the update with the reformulated gradient from \cref{eq: gradient of joint train} is as follows
\begin{equation}
\begin{split}
\hat\theta_{\mathrm{sha}} &=
\theta_{\mathrm{sha}} - \nabla_{\theta_{\mathrm{sha}}}\loss(\Theta) \\
&=\theta_{\mathrm{sha}} - M\frac{1}{T} \sum_{t=1}^{T}\nabla_{\theta_{\mathrm{sha}}}\loss_{t}(\theta_{\mathrm{sha}}, \theta_{\mathrm{sep}}^{t}).
\end{split}
\end{equation}
\vspace{-0.3cm}
\paragraph{Combination with gradient manipulation methods}
The application of sparse training can be seamlessly and effectively extended to improve various gradient manipulation methods in MTL. The update with the reformulated gradient from \cref{eq: gradient manipulation gradient} is as follows
\begin{equation}
\begin{split}
\hat\theta_{\mathrm{sha}} &=
\theta_{\mathrm{sha}} -
\nabla_{\theta_{\mathrm{sha}}}\loss_{\mathrm{gm}}(\Theta) \\
&=\theta_{\mathrm{sha}}-M\frac{1}{T} \sum_{t=1}^{T}\boldsymbol{w}_t\nabla_{\theta_{\mathrm{sha}}}\loss_{t}(\theta_{\mathrm{sha}}, \theta_{\mathrm{sep}}^{t}).
\end{split}
\end{equation}
\subsection{Theoretical analysis for sparse training}
After introducing sparse training into MTL, the optimization objective in \cref{eq:multi-objective} can be formed:
\begin{equation}
    \small
    \Theta^{*} = \arg\min_\Theta \loss(\Theta),
    s.t.\ \ \|(I-M)(\theta_{\mathrm{sha}}-\theta_{\mathrm{sha}}^{\mathrm{in}})\|^2=0,
    \label{eq: mtl s.t. M}
\end{equation}
where $\theta_\mathrm{sha}^{\mathrm{in}}$ is the initialized original model for $\theta$ and $I$ is identity matrix. According to Lagrangian duality, \cref{eq: mtl s.t. M} can be reformulated as:
\begin{equation}
  L=\min_\Theta \max_\lambda \mathcal{L}(\Theta)+\lambda\|(I-M)(\theta_{\mathrm{sha}}-\theta_\mathrm{sha}^{\mathrm{in}})\|^2.
\end{equation}
This can be transformed to optimize the upper bound $L$ of regularized problem:
\begin{equation}
  L_r= \min_\Theta \mathcal{L}(\Theta)+\|(I-M)(\theta_{\mathrm{sha}}-\theta_\mathrm{sha}^{\mathrm{in}})\|^2 \le L.
  \label{eq: regularized}
\end{equation}
Please see the supplemental material for proof. \citet{fu2023effectiveness} demonstrates that \cref{eq: regularized} has better stability and smaller generalization bound than only optimizing \cref{eq:multi-objective}, resulting in better performance.

\subsection{Parameter selection per neuron (PSN)}
\label{sec: neuron-based selection}
Several promising sparse training methods exist for single-task learning, but they are either time-consuming, requiring mask updates at each iteration \cite{sanh2020movement, mostafa2019parameter, xu2021raise}, or memory-intensive due to gradient calculations for all parameters \cite{zhang2023gradient, fu2023effectiveness}. In MTL, where multiple tasks are trained simultaneously, time efficiency is crucial. Thus, we adopt a one-time selection method, choosing parameters before training and keeping the mask fixed throughout. We consider the following two aspects for selection, magnitude of the parameter and involvement of all neurons in the network.
\vspace{-0.3cm}
\paragraph{The magnitude of parameters} 
Several studies have focused on model compression through the elimination of parameters with lower magnitudes \cite{han2015learning, frankle2018lottery}. This highlights the significance of parameters with larger magnitudes in neural networks, which is consistent with our experimental findings (See \cref{fig: different sparse method starplot}). 
The intuition behind this phenomenon lies in the fact that parameters with larger magnitudes exert a greater influence on altering neuron activation states through the activation function, wherein a neuron becomes active once the input surpasses a predefined threshold. Therefore, we exclusively select parameters with the highest magnitude for training multiple tasks.
\vspace{-0.3cm}
\paragraph{The involvement of all neurons}
A simple idea is to select a certain proportion of parameters with the highest magnitude from the neural network (NN), but this may prevent some neurons from being engaged during training and hinder effective model training due to the dependence of the NN state on neuron activation. Motivated by studies highlighting distinct roles for different components in NN \cite{wang2021pac, zhang2023gradient,Fan2020OnIO}, we posit that engaging all neurons is crucial for effective model training.
The rationale is that each neuron within the network possesses the inherent capability to finely adjust its activation state, thereby effectively adapting the overall NN state to the tasks, especially for learning multiple tasks simultaneously.
Our experiments further substantiate this assertion, as shown in \cref{fig: different sparse method starplot}. 

\begin{figure}[tb]
    \centering
        \includegraphics[width=0.5\linewidth]{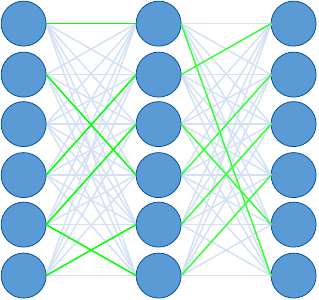}
        \caption{
        PSN. Top-1 highest-magnitude parameter among all input connections of each neuron is selected.
        }
        \label{fig: psn}
\end{figure}
\vspace{-0.3cm}
\paragraph{PSN} By integrating the two aspects, we select the top-K connections (weight/parameters) with the highest magnitude among all input connections for each neuron in the network (Please see \cref{fig: psn} for top-1 example). This approach facilitates the training process for fitting tasks by ensuring that every neuron possesses activation potential, while parameters with higher magnitudes facilitate easier activation of neurons. In this paper, sparse training refers to using this method to select parameters and training the selected parameter, unless otherwise specified.

\section{Experiments}
\label{sec: experiments}
Our experiments are conducted on comprehensive MTL benchmarks to evaluate the effectiveness of sparse training. First, we investigate if sparse training reduces gradient conflict. Subsequently, we examine its impact on performance across various MTL setups. The more details of the experiment are provided in \cref{sec: Detailed experiment setting}.
\subsection{EXPERIMENTAL SETUP}
\label{sec: exp: setup}
\paragraph{Dateset}
Our MTL datasets are categorized into three groups:
\romannumeral1) \textit{Dense prediction tasks}:
\textbf{\textit{NYUv2}} \cite{couprie2013indoor}: An indoor scene understanding dataset containing 1449 RGBD images with per-pixel labels across 13 classes, including semantic segmentation, depth estimation, and surface normal prediction.
\textbf{\textit{CityScapes}} \cite{cordts2016cityscapes}: 5000 street-view RGBD images with per-pixel annotations for 7-class semantic segmentation and depth estimation.
\romannumeral2) \textit{Multiple binary-classification tasks}:
\textbf{\textit{CelebA}} \cite{liu2015deep}: 200,000 facial images of 10,000 celebrities, each with 40 binary attributes for facial features. We use the first 10 attributes for 10 binary classification tasks due to limited computation.
\romannumeral3) \textit{Multiple multi-class classification tasks}:
\textbf{\textit{VTAB}} \cite{zhai2019large}: Containing 24 image understanding tasks with 1000 training examples per task. We use four tasks from it to create two multi-task benchmarks:
\textbf{\textit{Clevr}}: Simple 3D shapes with counting and depth prediction tasks.
\textbf{\textit{SmallNORB}}: Artificial objects with object azimuth and camera elevation prediction tasks.

\vspace{-0.3cm}
\paragraph{Baseline}
We evaluate our approach using various baselines including \romannumeral1) single-task learning (\textbf{\textit{STL}}): Each task is trained independently; \romannumeral2) \textbf{\textit{Joint Train}}: Training all tasks with average task loss; and 6 gradient manipulation methods including 3 direct and 3 indirect modification techniques. The former includes: \romannumeral3) \textbf{\textit{PCGrad}}: Projecting each task gradient onto the normal plane of other tasks \cite{PCGrad}; \romannumeral4) \textbf{\textit{CAGrad}}: Enhancing the optimization of average loss by explicitly regulating the minimum decrease across tasks \cite{CAGrad}; and \romannumeral5) \textbf{\textit{GradDrop}}: Stochastically dropping specific dimensions of the gradients based on their level of conflict. The latter includes \romannumeral6) \textbf{\textit{MGDA}}: Identifying the same descent direction for each task \cite{MGDA}; \romannumeral7) \textbf{\textit{IMTL-G}}: Determining the update direction by ensuring equal projections on gradients \cite{imtlg}; \romannumeral8) \textbf{\textit{NashMTL}}: Treating MTL as a bargaining game to optimize all tasks \cite{nash}.

\vspace{-0.3cm}
\paragraph{Model} We experiment with several architectures including: \romannumeral1) \textit{CNN-based}: \textbf{\textit{MTAN}} \cite{SegNet_attention} incorporates an attention mechanism into the SegNet \cite{badrinarayanan2017segnet}.
\romannumeral2) \textit{Transformer-based}. \textbf{\textit{SAM}} \cite{kirillov2023segment} is a strong visual foundation model for segmentation. \textbf{\textit{\textbf{\textit{ViT-B/16}}}} \cite{dosovitskiy2020image} and \textbf{\textit{Swin Transformer}} \cite{liu2021swin} are vision classification models pre-trained on ImageNet21K \cite{deng2009imagenet}. All experiments were conducted on pre-trained SAM, ViT and Swin (except for randomly initialized MTAN), unless otherwise specified.

\vspace{-0.3cm}
\paragraph{Evaluation} \romannumeral1) \textbf{\textit{Relative task drop}} ($\bm{\Delta}\bm{m} \%$). Following \cite{maninis2019attentive}, we evaluate the MTL overall performance for a baseline $b$ by computing the average performance drop against STL $s$ over $\{\mathcal{T}_t\}_{t=1}^{T}$ tasks and $K_{\mathcal{T}_t}$ metrics for each $\mathcal{T}_t$: $\Delta m\% = (\frac{1}{T}\sum_{t=1}^{T} \frac{1}{K_{\mathcal{T}_t}} \sum_{k=1}^{K_{\mathcal{T}_t}} (-1)^{\delta_k} (M_{b}^{k} - M_{s}^{k}) / M_{s}^{k}) \times 100$  where $M_{b}^{k}$, $M_{s}^{k}$ are the value of metrics $k$ evaluated with $b$ and $s$ respectively. $\delta_k = 1$ if the $M^k$ is higher the better and 0 otherwise.
\romannumeral2) \textbf{\textit{Average incidence of GC}} ($\bm{p}\%$). We evaluate the extent of gradient conflict for a baseline by calculating the average incidence of GC over epochs during training. Given $T$ tasks, $E$ epochs, and $I$ iterations per epoch,  $\bm{p} \% = \frac{1}{EI}\sum_{e=1}^{E}\sum_{i=1}^{I} (N_{gc} / N_{all}) \times 100$, where $N_{gc}$ and $N_{all}$ represent the number of occurrence of gradient conflicts between two tasks for all task combinations $\binom{T}{2}$ and the number of the combinations in each iteration during training, respectively.

\subsection{Incidence of gradient conflict}
\label{sec: exp: Occurrence of gradient conflict}

We train a MTL model using the \textit{Joint Train} and 6 state-of-the-art gradient manipulation techniques including \textit{PCGrad}, \textit{CAGrad}, \textit{GradDrop}, \textit{MGDA}, \textit{IMTL-G} and \textit{NashMTL} and then introduce our 
sparse training strategy to these methods. Throughout the training process, we record instances of GC between any two tasks among all tasks for each training iteration and then calculate the average incidence of GC both over all epochs and the last 50\% epochs. The observations of the SAM model on the NYU-v2 dataset are provided below. Similar results on other datasets and models are shown in \cref{sec: supp NYU-v2 on MTAN}, \cref{sec: supp CelebA on Swin}, \cref{sec: supp CityScapes on MTAN} and \cref{sec: supp SmallNORB on ViT}.

\vspace{-0.3cm}
\paragraph{Gradient manipulation methods cannot effectively reduce the incidence of gradient conflict} 
The gradient manipulation methods ~\citep{PCGrad,Graddrop,CAGrad,nash,liu2023famo,imtlg,MGDA} aim to modify conflicting gradients that are prevalent during the joint training of MTL. As shown in \cref{tab:gc of different methods}, the average incidence of GC using \textit{Joint train} is 31.89\% across all training epochs and 35.85\% over the last 50\% epochs. The incidence of GC cannot be effectively reduced by any gradient magnitude methods compared with the \textit{Joint train}, as shown in \cref{fig:gc of different methods} and \cref{tab:gc of different methods}.  The reason is that these methods can only make the conflicting gradients not conflict when the GC occurs, rather than proactively prevent the occurrence of GC. The incidence of GC is even exacerbated by these methods, particularly MGDA showing a significant increase of 8.55\% compared to \textit{Joint Train}. Notably, these findings are consistent with \citep{shi2023recon}, where they provide the distribution of the angles between the two task gradients.

\begin{figure}[t]
        \subfloat[Joint Train]{
          \centering
        \includegraphics[width=0.48\linewidth]{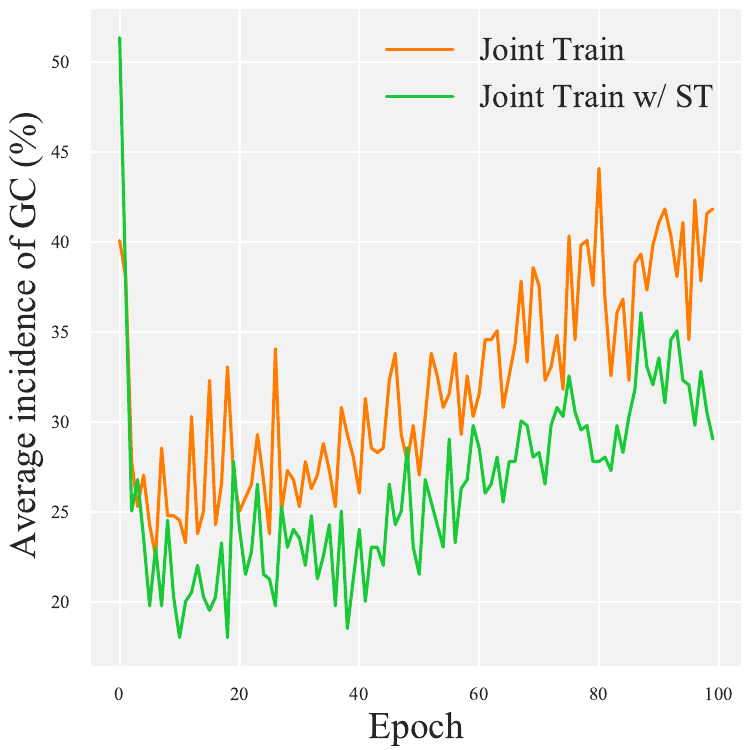}
            \label{fig:joint}
        }
        \subfloat[PCGrad]{
        \centering
        \includegraphics[width=0.48\linewidth]{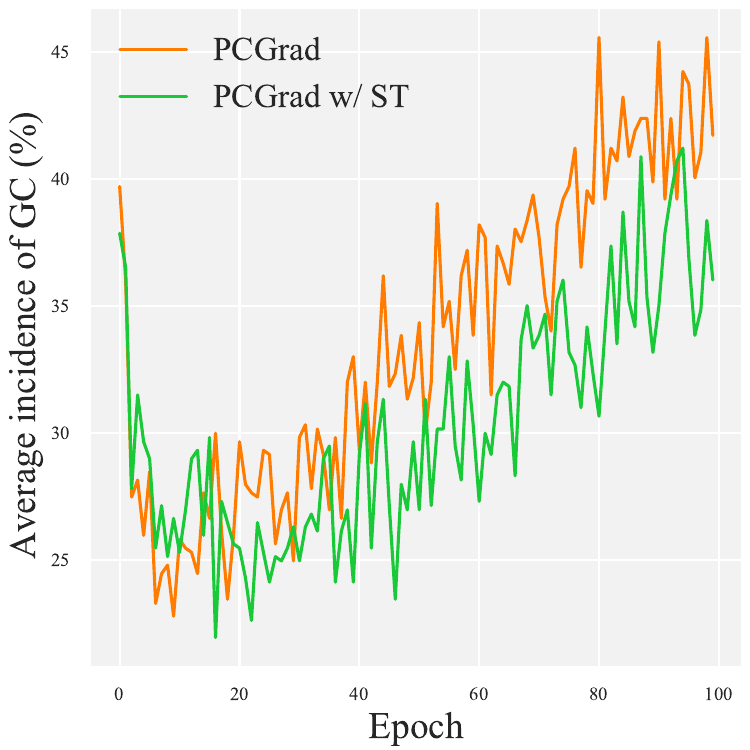}
            \label{fig:PCGrad}
        }
        \caption{The incidence of GC between tasks during training SAM on NYUv2 dataset. The top and bottom figures are \textit{Joint Train} and \textit{PCGrad} respectively. Please see \cref{fig:number of gc} in \cref{sec: supp NYU-v2 on SAM} for more results on other gradient manipulation methods.
        }
        \label{fig:incidence of gc (part)}
\end{figure}

\vspace{-0.3cm}
\paragraph{Sparse training effectively decreases the occurrence of gradient conflict} 
As shown in \cref{tab:gc of different methods}, after combining sparse training with all methods, including \textit{Joint Train} and gradient manipulation methods, the average incidence of gradient conflict is effectively reduced over all epochs. For example, ST in \textit{Joint Train} reduced the incidence over all epochs by 5.56\%. The phenomenon of gradient conflict reduction is consistently observed in nearly every training epoch, as illustrated in \cref{fig:incidence of gc (part)}, which further demonstrates the effectiveness of ST for decreasing gradient conflict.
In addition, all methods with ST exhibit a greater improvement in the average incidence of gradient conflict during the last 50\% epochs compared to all epochs, which implies a greater level of prevention of gradient conflict with the progress of sparse training.
For instance of NashMTL, there is a threefold improvement in the average incidence of gradient conflict during the last 50\% epochs compared to all epochs.


\begin{table}[t] 
    \centering
    \resizebox{0.4\textwidth}{!}{
    \begin{tabular}{ccll}
    \toprule 
     & \multirow{2}{*}{\textbf{Methods}}& \multicolumn{2}{c}{Average incidence of GC ($\%$)}\\
     \cmidrule(lr){3-4} &                                  &All epochs& Last 50\% epochs\\
    \toprule
    &Joint Train& 31.89&35.85 \\
    &     w/ ST & 26.33 (\textcolor{blue}{5.56}) &29.14 (\textcolor{blue}{6.71}) \\
    \midrule
     & PCGrad & 33.69& 38.70\\
     & w/ ST & 30.33 (\textcolor{blue}{3.36})&33.46 (\textcolor{blue}{5.24})\\
    \midrule
     & CAGrad& 34.26& 39.97\\
     & w/ ST& 31.50 (\textcolor{blue}{2.76})& 34.68 (\textcolor{blue}{5.29}) \\
    \midrule
     & GradDrop &33.56 &38.45 \\
     & w/ ST & 30.95 (\textcolor{blue}{2.61})& 33.93 (\textcolor{blue}{4.52})\\
    \midrule
     & MGDA & 40.44& 44.77\\
     & w/ ST& 40.05 (\textcolor{blue}{0.39})&42.34 (\textcolor{blue}{2.43})\\
    \midrule
     & IMTL-G & 32.15&37.13 \\
     & w/ ST & 28.45 (\textcolor{blue}{3.70})&31.34 (\textcolor{blue}{5.79})\\
    \midrule
     & NashMTL & 36.67& 39.58\\
     & w/ ST & 35.51 (\textcolor{blue}{1.16})& 35.48 (\textcolor{blue}{4.10})\\
    \bottomrule
    \end{tabular}}
    \caption{Average incidence of GC between tasks for different methods. We compute the average incidence of GC over all epochs and the last 50\% epochs during training SAM on NYUv2. The improvement by sparse training is provided in (\textcolor{blue}{$\bullet$}).} 
    \label{tab:gc of different methods}
\end{table}

\subsection{Performance on diverse benchmarks}
\label{sec: Performance on diverse benchmarks}
It is natural to investigate whether reducing gradient conflict during training through sparsity can enhance performance on common benchmarks. In this section, we present diverse benchmarks to demonstrate the effectiveness of ST.

\begin{table*}[t]
\centering
\resizebox{0.9\textwidth}{!}{
\begin{tabular}{cccccccccccccccccc}
\toprule 
 & \multirow{3}{*}{\textbf{Methods}}  & \multicolumn{2}{c}{Segmentation}  & \multicolumn{2}{c}{Depth}  & \multicolumn{5}{c}{Surface Normal}  &  \multirow{3}{*}{$\mathbf{\Delta m \%} \downarrow$} & \\
 \cmidrule(lr){3-4} \cmidrule(lr){5-6} \cmidrule(lr){7-11}
 &  & \multirow{2}{*}{mIoU $\uparrow$} & \multirow{2}{*}{Pix Acc $\uparrow$}  & \multirow{2}{*}{Abs Err $\downarrow$} & \multirow{2}{*}{Rel Err $\downarrow$}  & \multicolumn{2}{c}{Angle Distance $\downarrow$} & \multicolumn{3}{c}{Within $t^\circ$  $\uparrow$} &   \\
 \cmidrule(lr){7-8} \cmidrule(lr){9-11}
 &  &  &  &  &  &  Mean & Median  & 11.25   & 22.5   & 30 &   \\
\toprule
  & \multicolumn{1}{c}{STL} & $58.62$& $79.20$&$0.3810$ &$0.1553$ &$19.29$ &$12.64$ &  $46.37$& $72.19$& $80.73$& $-$\\
  \midrule
 & \multicolumn{1}{c}{Joint Train} & $59.09$& $79.61$&$0.3348$ &$0.1360$ &$22.34$ &$16.33$  & $35.46$& $64.02$& $75.20$ &$6.763$\\
& \multicolumn{1}{c}{w/ ST} & \cellcolor[HTML]{CCFFCC}$60.03$& \cellcolor[HTML]{CCFFCC}$79.96$& \cellcolor[HTML]{CCFFCC}$0.3320$&\cellcolor[HTML]{CCFFCC}$0.1353$ & \cellcolor[HTML]{CCFFCC}$21.98$& \cellcolor[HTML]{CCFFCC}$15.92$ & \cellcolor[HTML]{CCFFCC}$36.69$&\cellcolor[HTML]{CCFFCC}$64.92$ &\cellcolor[HTML]{CCFFCC}$75.82$  &\cellcolor[HTML]{CCFFCC}$5.314$ \\
\midrule
 & \multicolumn{1}{c}{PCGrad} & $59.18$& $80.12$&$0.3258$ &$0.1323$ &$21.81$ &$15.72$  & $36.92$& $65.49$& $76.26$ &$4.584$\\
& \multicolumn{1}{c}{w/ ST} &\cellcolor[HTML]{CCFFCC}$59.37$&\cellcolor[HTML]{CCFFCC}$\textbf{80.33}$& $0.3272$&$0.1330$ &\cellcolor[HTML]{CCFFCC}$21.53$&\cellcolor[HTML]{CCFFCC}$15.39$ & \cellcolor[HTML]{CCFFCC}$38.02$&\cellcolor[HTML]{CCFFCC}$66.09$ &\cellcolor[HTML]{CCFFCC}$76.71$ &\cellcolor[HTML]{CCFFCC}$3.741$ \\
\midrule
 & \multicolumn{1}{c}{CAGrad} & $59.78$ & $80.16$ & $\textbf{0.3215}$ & $\textbf{0.1305}$ & $19.92$ & $13.40$ & $43.87$ & $70.47$ & $79.59$ & $-1.816$\\ 
 & \multicolumn{1}{c}{w/ ST} & \cellcolor[HTML]{CCFFCC}$60.33$ &\cellcolor[HTML]{CCFFCC}$80.20$ & $0.3232$ & $0.1306$ & \cellcolor[HTML]{CCFFCC}$19.74$ &\cellcolor[HTML]{CCFFCC}$13.20$ & \cellcolor[HTML]{CCFFCC}$44.42$ &\cellcolor[HTML]{CCFFCC}$71.04$ & \cellcolor[HTML]{CCFFCC}$80.02$ &\cellcolor[HTML]{CCFFCC}$\bold{-2.423}$\\
\midrule
 & \multicolumn{1}{c}{GradDrop} & $59.02$ & $79.80$ & $0.3283$ & $0.1321$ & $22.03$ & $15.95$ & $36.42$ & $64.90$ & $75.79$ & $5.323$\\
 & \multicolumn{1}{c}{w/ ST} &\cellcolor[HTML]{CCFFCC}$59.74$ & \cellcolor[HTML]{CCFFCC}$80.32$ &\cellcolor[HTML]{CCFFCC}$0.3278$ & $0.1322$ & \cellcolor[HTML]{CCFFCC}$21.81$ &\cellcolor[HTML]{CCFFCC}$15.63$  &\cellcolor[HTML]{CCFFCC}$37.40$ & \cellcolor[HTML]{CCFFCC}$65.50$ &\cellcolor[HTML]{CCFFCC}$76.15$ & \cellcolor[HTML]{CCFFCC}$4.329$\\
\midrule
 & \multicolumn{1}{c}{MGDA} & $37.43$ & $67.58$ & $0.4427$ & $0.1810$ & $19.23$ & $\textbf{12.61}$  & $46.43$ & $\textbf{72.35}$ & $\textbf{80.87}$ &$9.162$\\
 & \multicolumn{1}{c}{w/ ST}& \cellcolor[HTML]{CCFFCC}$41.60$ & \cellcolor[HTML]{CCFFCC}$69.96$ & \cellcolor[HTML]{CCFFCC}$0.4414$ & \cellcolor[HTML]{CCFFCC}$0.1778$ & \cellcolor[HTML]{CCFFCC}$\textbf{19.22}$ & $\textbf{12.61}$ & \cellcolor[HTML]{CCFFCC}$\textbf{46.44}$ & $72.29$ & $80.80$ &\cellcolor[HTML]{CCFFCC}$7.791$\\
\midrule
 & \multicolumn{1}{c}{IMTL-G} & $\textbf{60.64}$ & $80.29$ & $0.3324$ & $0.1348$ & $19.85$ & $13.37$ &  $43.92$ & $70.78$ & $79.9$ & $-1.537$\\
 & \multicolumn{1}{c}{w/ ST} & $60.35$ & $80.18$ & $0.3347$ & $0.1350$ & \cellcolor[HTML]{CCFFCC}$19.65$ & \cellcolor[HTML]{CCFFCC}$13.15$  & \cellcolor[HTML]{CCFFCC}$44.66$ & \cellcolor[HTML]{CCFFCC}$71.25$ & \cellcolor[HTML]{CCFFCC}$80.20$ & \cellcolor[HTML]{CCFFCC}$-1.955$\\
\midrule
 & \multicolumn{1}{c}{NashMTL} & $59.42$ &$80.20$ & $0.3303$ & $0.1341$ & $19.90$ & $13.39$ &  $43.86$ & $70.65$ & $79.72$ & $-1.295$\\
 & \multicolumn{1}{c}{w/ ST} & $59.36$ & $79.98$ &\cellcolor[HTML]{CCFFCC}$0.3278$ & \cellcolor[HTML]{CCFFCC}$0.1323$ & \cellcolor[HTML]{CCFFCC}$19.63$ & \cellcolor[HTML]{CCFFCC}$13.02$  &\cellcolor[HTML]{CCFFCC}$45.06$ &\cellcolor[HTML]{CCFFCC}$71.31$ & \cellcolor[HTML]{CCFFCC}$80.15$ &\cellcolor[HTML]{CCFFCC}$-2.384$\\
\bottomrule
\end{tabular}
}
\caption{The test performance on NYU-v2 dataset training on SAM model. The green cell color indicates that sparse training improves the performance of joint training or gradient manipulation methods. The best result is highlighted in bold.}
\label{tab:nyu sam}
\end{table*}

\begin{table*}[tb]
    \centering
    \resizebox{0.7\textwidth}{!}{
    \begin{tabular}{cccccccccc}
    \toprule 
     &  &  \multicolumn{2}{c}{\textbf{CelebA}} & \multicolumn{3}{c}{\textbf{Clevr}} &  \multicolumn{1}{c}{\textbf{SmallNORB}} & \multicolumn{1}{c}{\textbf{NYU-v2}} & \multicolumn{1}{c}{\textbf{CityScapes}}\\
     \cmidrule(lr){3-4} \cmidrule(lr){5-7} \cmidrule(lr){8-8} \cmidrule(lr){9-9} \cmidrule(lr){10-10} 
     & \multirow{1}{*}{\textbf{Methods}}& & \multirow{1}{*}{$\mathbf{\Delta m \%} \downarrow$} &\multicolumn{1}{c}{Counting}  & \multicolumn{1}{c}{Depth} & \multirow{2}{*}{$\mathbf{\Delta m \%} \downarrow$}& \multirow{2}{*}{$\mathbf{\Delta m \%} \downarrow$}& \multirow{2}{*}{$\mathbf{\Delta m \%} \downarrow$}& \multirow{2}{*}{$\mathbf{\Delta m \%} \downarrow$}\\
      \cmidrule(lr){5-5} \cmidrule(lr){6-6} 
     &  &  & (F1) & \multicolumn{1}{c}{(Top 1 $\uparrow$)}  &   \multicolumn{1}{c}{(Top 1 $\uparrow$)} & \\
    \toprule
    & \multicolumn{2}{c}{STL} & $-$& $58.64$  &$57.68$ & $-$ & $-$ & $-$& $-$\\
    \midrule
    & \multicolumn{2}{c}{Joint Train} & $3.12$ &$54.86$  & $54.68$ &  $5.84$&10.70 &5.59&26.87 \\
    & \multicolumn{2}{c}{w/ ST } &\cellcolor[HTML]{CCFFCC}$2.03$& \cellcolor[HTML]{CCFFCC}$\textbf{61.80}$ & \cellcolor[HTML]{CCFFCC}$54.81$ & \cellcolor[HTML]{CCFFCC}$-0.21$ & \cellcolor[HTML]{CCFFCC}10.11&\cellcolor[HTML]{CCFFCC}2.49&\cellcolor[HTML]{CCFFCC}17.48\\
    \midrule
    & \multicolumn{2}{c}{ PCGrad} &$1.70$ & $49.01$&  $53.39$ & $11.93$&9.99 & 3.97 &19.96\\
    & \multicolumn{2}{c}{w/ ST} &\cellcolor[HTML]{CCFFCC}$1.42$&\cellcolor[HTML]{CCFFCC}$59.01$   & \cellcolor[HTML]{CCFFCC}$55.29$ &  $\cellcolor[HTML]{CCFFCC}1.75$& \cellcolor[HTML]{CCFFCC}9.71&\cellcolor[HTML]{CCFFCC}1.98&\cellcolor[HTML]{CCFFCC}19.22\\
    \midrule
    & \multicolumn{2}{c}{CAGrad } &$1.96$& $49.33$&   $53.67$ & $11.41$ &10.50 & 0.20&16.26\\
    & \multicolumn{2}{c}{w/ ST} &\cellcolor[HTML]{CCFFCC}$1.23$&\cellcolor[HTML]{CCFFCC}$58.51$   & \cellcolor[HTML]{CCFFCC}$55.27$ &  \cellcolor[HTML]{CCFFCC}$2.19$&\cellcolor[HTML]{CCFFCC}10.22 &\cellcolor[HTML]{CCFFCC}-2.76&\cellcolor[HTML]{CCFFCC}8.88\\
    \midrule
    & \multicolumn{2}{c}{GradDrop } &$1.18$& $49.02$ &  $52.88$ & $12.36$&11.73 &3.58&20.34 \\
    & \multicolumn{2}{c}{w/ ST} &\cellcolor[HTML]{CCFFCC}$0.83$&\cellcolor[HTML]{CCFFCC}$58.87$  &  \cellcolor[HTML]{CCFFCC}$54.07$&   \cellcolor[HTML]{CCFFCC}$2.94$&\cellcolor[HTML]{CCFFCC}10.76 &\cellcolor[HTML]{CCFFCC}1.38&\cellcolor[HTML]{CCFFCC}17.45\\
    \midrule
    & \multicolumn{2}{c}{MGDA} &$-0.41$& $49.56$&   $55.97$ & $9.22$& 10.15&1.38 &6.91\\
    & \multicolumn{2}{c}{w/ ST} &\cellcolor[HTML]{CCFFCC}$\textbf{-1.08}$&\cellcolor[HTML]{CCFFCC}$58.23$  &  \cellcolor[HTML]{CCFFCC}$\textbf{56.91}$ &  \cellcolor[HTML]{CCFFCC}$1.02$&\cellcolor[HTML]{CCFFCC}9.79 &\cellcolor[HTML]{CCFFCC}-3.18&\cellcolor[HTML]{CCFFCC}\textbf{3.17}\\
    \midrule
    & \multicolumn{2}{c}{IMTL-G } &$0.97$& $54.99$& $54.51$ & $5.87$&10.19 &-0.76 &10.65\\
    & \multicolumn{2}{c}{w/ ST} &\cellcolor[HTML]{CCFFCC}$0.19$&\cellcolor[HTML]{CCFFCC}$61.05$   & \cellcolor[HTML]{CCFFCC}$56.73$&  \cellcolor[HTML]{CCFFCC}$\textbf{-1.24}$&\cellcolor[HTML]{CCFFCC}10.15 &\cellcolor[HTML]{CCFFCC}-3.18&\cellcolor[HTML]{CCFFCC}7.10\\
    \midrule
    & \multicolumn{2}{c}{NashMTL } &$3.59$& $47.04$  & $53.07$ & $13.89$&10.84 & -4.04&6.68\\
    & \multicolumn{2}{c}{w/ ST} &\cellcolor[HTML]{CCFFCC}$3.22$&\cellcolor[HTML]{CCFFCC}$58.61$ & \cellcolor[HTML]{CCFFCC}$54.97$&   \cellcolor[HTML]{CCFFCC}$2.37$& \cellcolor[HTML]{CCFFCC}\textbf{9.57}&\cellcolor[HTML]{CCFFCC}\textbf{-5.11}&\cellcolor[HTML]{CCFFCC}3.99\\
    \bottomrule
    \end{tabular}
    }
    \caption{The test performance on CelebA, Clevr, SmallNORB, NYU-v2 and CityScapes dataset. CelebA is trained on Swin Transformer. Clevr and SmallNORB are trained on ViT. NYU-v2 and CityScapes are trained on MTAN. We only present $\Delta m \%$ for limited space. Please see \cref{tab:nyu mtan}, \cref{tab:city mtan} and \cref{tab:smallnorb} for detailed results in supplemental materials. The green cell color indicates that sparse training improves the performance of joint training or gradient manipulation methods. The best result is highlighted in bold. }
     \label{tab:celeba clever SmallNORB nyuv2 city}
\end{table*}

\begin{figure*}[tb]
\centering
\subfloat[Different model size]{
    \includegraphics[width=0.32\linewidth]{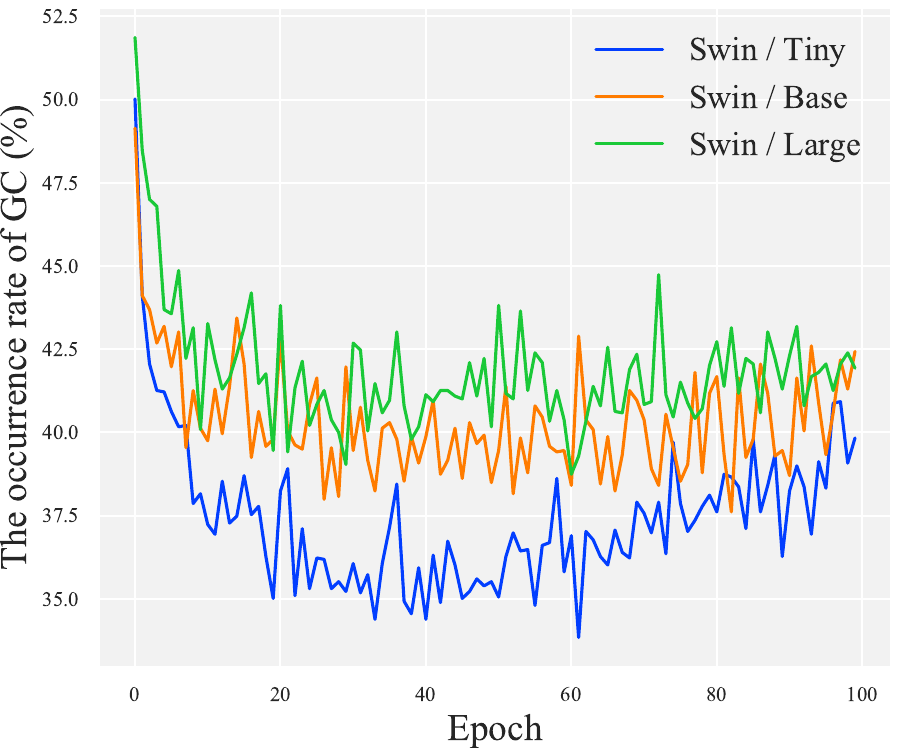}
    \label{fig: gc of different swin model size}
}
\hfill
\centering
\subfloat[Different num. of parameters]{
    \includegraphics[width=0.32\linewidth]{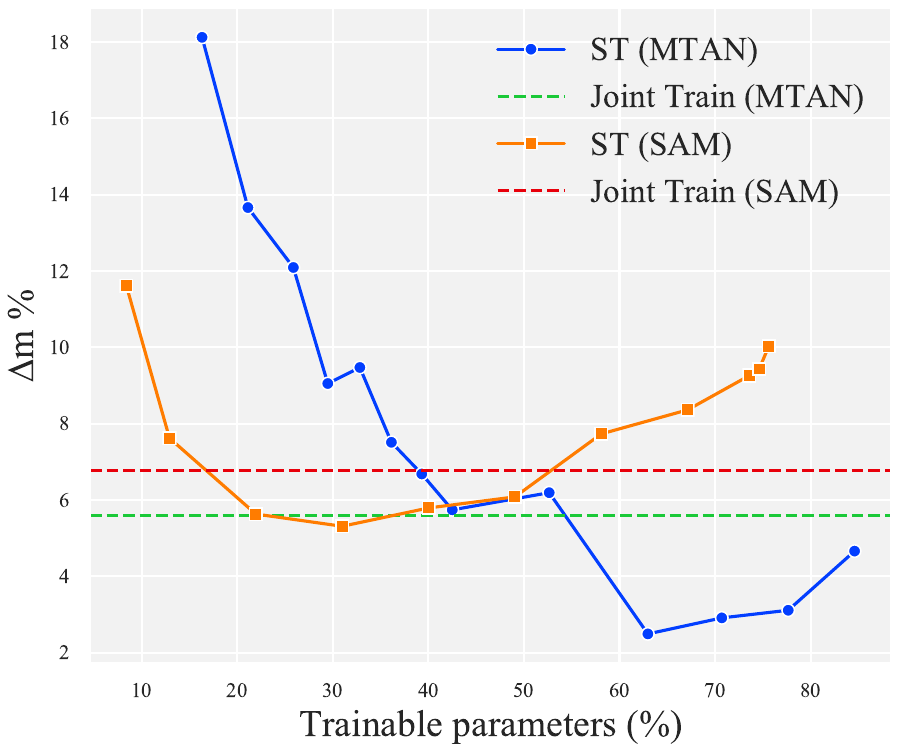}
    \label{fig: different tunable number}
}
\hfill
\centering
\subfloat[Different sparse methods]{
    \includegraphics[width=0.27\linewidth]{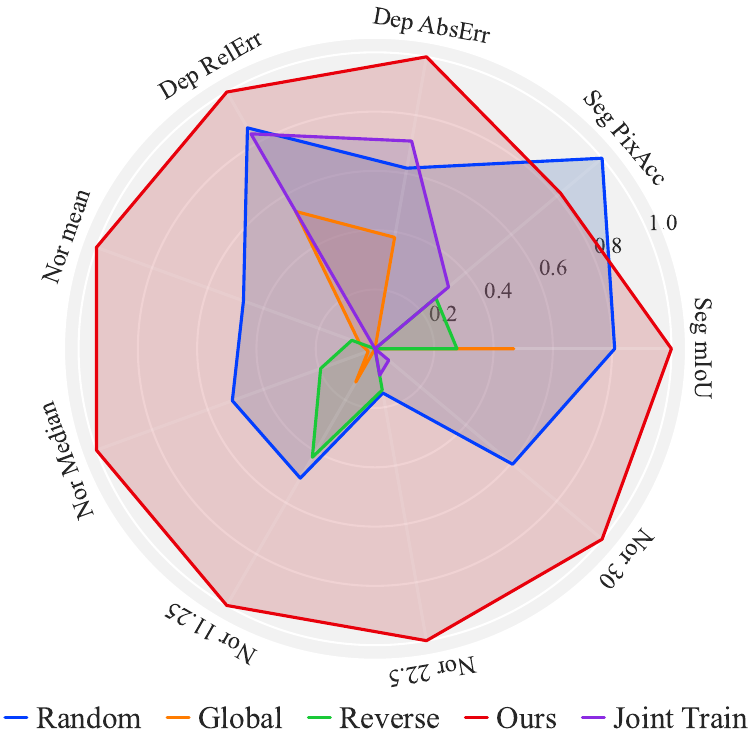}
    \label{fig: different sparse method starplot}
}
\caption{Ablation study for \textit{Joint Train} with NYU-v2 dataset. (a) The average incidence of GC during joint training on different sizes of Swin transformers. Please see the numerical statics for all epochs in \cref{tab:gc of different swin model size} in \cref{sec: supp NYU-v2 on Swin}. (b) The different number of trainable parameters for MTAN and SAM models. (C) Different sparse methods training on SAM. Metrics for all tasks are min-max normalized. Please see \cref{tab:ablation (full): different sparse training methods} for detailed results in \cref{sec: supp: Ablation study}.}
\vspace{-0.1cm}
\label{fig:ablation: model size, sparse method and tunable number}
\end{figure*}

\vspace{-0.3cm}
\paragraph{Sparse training improves the performance for all state-of-the-art methods} 
The performance of \textit{Joint Train} and all gradient manipulation methods is consistently improved by sparse training, as demonstrated in \cref{tab:nyu sam} for NYU-v2 benchmarks. Specifically, sparse training not only enhances overall task performance but also improves individual task performance for the majority of methods. For example, in \cref{tab:nyu sam}, \textit{Joint Train} demonstrates improvements across all individual tasks through sparse training. Similarly, as shown in \cref{tab:celeba clever SmallNORB nyuv2 city}, all methods exhibit notable improvements by sparse training on  CelebA, Clevr, SmallNORB and CityScapes benchmarks.
\vspace{-0.3cm}
\paragraph{Effectiveness on both pre-trained and randomly initialized models}
Our study primarily focuses on the sparse training for large pre-trained models, because leveraging prior knowledge from these models can be beneficial for MTL and our experimental results demonstrate that larger models exhibit a more severe gradient conflict, as shown in \cref{fig: gc of different swin model size}. However, in order to ensure a fair comparison with related works that manipulate gradients in small and randomly initialized models, we also conduct experiments under the same setting as theirs to further demonstrate the effectiveness of sparse training. As shown in \cref{tab:celeba clever SmallNORB nyuv2 city}, we observe that even for the small randomly initialized models, the performance of joint training and all gradient manipulation methods is improved by sparse training. Please see  \cref{tab:nyu mtan} and \cref{tab:city mtan} for the detailed results in the Appendix.

\vspace{-0.3cm}
\paragraph{Generalization on different architectures and MTL tasks}
To evaluate the generalization across diverse architectures and MTL tasks, we conducted experiments on both CNN-based models and transformer-based models with varying visual MTL capabilities. Specifically, our MTL tasks encompassed visual classification (CelebA, Clevr and SmallNORB) and visual dense prediction (NYU-v2  and CityScapes). For the former, we utilized Swin Transformer and ViT as backbones for multiple binary classification tasks (\cref{tab:celeba clever SmallNORB nyuv2 city}) and two multi-class classification tasks (\cref{tab:celeba clever SmallNORB nyuv2 city}, and \cref{tab:smallnorb} in Appendix), respectively. The latter involved predicting dense masks for each task, necessitating an encoder-decoder structure to generate corresponding masks. We explored two types of structures: a symmetrical encoder-decoder structure with a CNN-based model, e.g. MTAN (\cref{tab:celeba clever SmallNORB nyuv2 city}, and \cref{tab:nyu mtan,tab:city mtan} in Appendix) and an asymmetric structure with a heavy-weight encoder and a light-weight decoder using a transformer-based model, e.g. SAM (\cref{tab:nyu sam} in Appendix). As shown in these tables, the efficacy of sparse training in improving all baselines across various architectures and MTL tasks underscores its robust generalization capability.






\subsection{Ablation study}

\paragraph{The larger the model, the more severe gradient conflicts.}
In this paper, we focus more on investigating the gradient conflict in the pre-trained large models as larger models demonstrated a more severe phenomenon of gradient conflict. This can be observed in \cref{fig: gc of different swin model size}, where \textit{Swin/Tiny} demonstrates significantly less gradient conflict compared to \textit{Swin/Base} and \textit{Swin/Large}. 
It is worth noting that although larger models tend to experience more severe gradient conflicts, this does not necessarily lead to inferior performance compared to smaller models with milder gradient conflicts. This discrepancy can be attributed to differences in model capacity and the prior knowledge embedded through pre-training.
Nevertheless, this observation underscores the importance of exploring methods to mitigate gradient conflicts in larger models. Within the same model architecture and size, reducing gradient conflicts has been shown to improve performance, as evidenced by works such as \cite{PCGrad, CAGrad}. Addressing severe gradient conflicts in larger models may thus unlock their full potential, enabling better utilization of their capacity and capabilities.

\vspace{-0.3cm}
\paragraph{Effortless search for the number of trainable parameters.} We explore the effect of trainable parameter numbers for ST.
The results in \cref{fig: different tunable number} show that the pre-trained model (SAM) and the randomly initialized model (MTAN) have different optimal trainable parameter numbers. MTAN requires $\sim$60\% of the parameters, while SAM needs only $\sim$30\%, leveraging information from the pre-trained model. In our paper, most of the experiments use these proportions for ST and achieve better results (please see \cref{table: Number of trainable parameters} in \cref{sec: Number of trainable parameters} for the detailed number). 
Additionally, ST offers a wide range of trainable parameter options that outperform \textit{Joint Train}, which implies that hyperparameter search for the number of trainable parameters becomes effortless. Specifically, both models have a $\sim$40\% probability of yielding superior outcomes.

\vspace{-0.3cm}
\paragraph{Effectiveness for both higher magnitude and neural-level selection.}
We investigate various parameter selection approaches:
\textit{Random}: Randomly selecting parameters from the network;
\textit{Global}: Choosing parameters with the highest magnitude from the whole network instead of the input connections of each neuron in the network (\textit{Ours});
\textit{Reverse}: Selecting parameters with the lowest magnitude among input connections of each neuron.
For a fair comparison, we maintain the same selected number.
The results in \cref{fig: different sparse method starplot} indicate that higher magnitude values are superior to lower ones (\textit{Ours} $>$ \textit{Reverse}). Furthermore, it is crucial to evenly select parameters from the entire network (\textit{Ours} $>$ \textit{Random} $>$ \textit{Global}), as  \textit{Ours} ensure that the parameters of input connection for each neuron are selected,  and \textit{Random} guarantees an equal proportion of parameters is selected in each block of the network, whereas this is not the case for \textit{Global} (see \cref{fig: selected number of parameters for diff sparse methods} for detailed statistics in Appendix).

\section{Conclusion}
In this paper, the occurrence of gradient conflict in multi-task learning is extensively investigated from a novel perspective: sparse training. Extensive experiments demonstrate that sparse training transferring high-dimensional space into low-dimensional space effectively reduces the incidence of gradient conflict during training while preserving the integrity of the original model. Furthermore, combining sparse training with other gradient manipulation methods significantly improves performance for multi-task learning.



\clearpage

{
    \small
    \bibliographystyle{ieeenat_fullname}
    \bibliography{main}
}
\appendix

\clearpage
\setcounter{page}{1}
\maketitlesupplementary

In this supplemental material, we provide extra details about the content in the main body of the paper.  First, we provide detailed proof for \Cref{eq: regularized} in \Cref{sec: supp proof}. Then, we discuss the limitations of our work in \Cref{sec: limitations}. Moreover, the broader impacts of our research are discussed in \Cref{sec: Broader Impacts}. In addition, we present all hyperparameters and experiment settings in \Cref{sec: Detailed experiment results} for a better understanding of the experiments and reproduction to the readers. We also provide the extended related works in \Cref{sec: Extended related work}. Finally, the additional experiment results are demonstrated in \Cref{sec: Detailed experiment results}, which further indicate the effectiveness of our proposed method and the consistency with the claim in the main body of the paper.

\section{Proof for \Cref{eq: regularized}}
\label{sec: supp proof}

    According to Lagrangian duality, \cref{eq: mtl s.t. M} can be reformulated as:
    $$\begin{aligned}
      L=&\min_\Theta \max_\lambda \mathcal{L}(\Theta)+\lambda\|(I-M)(\theta_{\mathrm{sha}}-\theta_\mathrm{sha}^{\mathrm{in}})\|^2\\
      \ge& \max_\lambda \min_\Theta  \mathcal{L}(\Theta)+\lambda\|(I-M)(\theta_{\mathrm{sha}}-\theta_\mathrm{sha}^{\mathrm{in}})\|^2\\
      \ge& \min_\Theta  \mathcal{L}(\Theta)+\|(I-M)(\theta_{\mathrm{sha}}-\theta_\mathrm{sha}^{\mathrm{in}})\|^2\\
  \end{aligned}$$
  where $\lambda$ is the Lagrangian multiplier.

\section{Limitations}
\label{sec: limitations}
Due to the limited computational resources, we employ grid searches in the \textit{Joint train} method to determine the optimal hyperparameter for the number of trainable parameters, which is then utilized across all gradient manipulation methods. However, it is possible that these methods may benefit from a more optimized hyperparameter selection for the number of trainable parameters. Furthermore, sparse training can effectively mitigate gradient conflicts between tasks in MTL by reducing the dimensionality of parameter space and limiting their impact on updates between tasks. The regularization constitutes one of the theory's reasons. Nevertheless, we anticipate that our future research will contribute to a deeper comprehension of multi-task learning and subsequently enhance the performance of MTL.

\section{Broader Impacts}
\label{sec: Broader Impacts}
The nature of our research does not directly contribute to societal impact; however, like any machine learning paper, it has the potential to adversely affect society through automation and job displacement. While it is challenging to predict specific risks, similar to any technology, inadequate regulation may lead to an exacerbation of social and economic inequality. The positive aspect lies in the potential environmental impact of our work, as multi-task learning enables information sharing among tasks, thereby reducing data requirements and further minimizing energy consumption during training.

\section{Detailed experiment setting}
\label{sec: Detailed experiment setting}

\subsection{Number of trainable parameters}
\label{sec: Number of trainable parameters}
We provide the number of trainable parameters for all experiments conducted in our paper. As shown in \Cref{table: Number of trainable parameters}, most of them have the same percentage of trainable parameters within a model across different methods. In addition, in general, we can observe that sparse training for the pre-trained model needs $\sim$30\% while that for random initialized model needs $\sim$60\%.

\begin{table*}[t]
\centering
\begin{tabular}{lcccccc}
\toprule
\multicolumn{1}{c}{}       & \multicolumn{4}{c}{Pre-trained model}       & \multicolumn{2}{l}{Random initialized model} \\ \cmidrule(lr){2-5}  \cmidrule(lr){6-7} 
\multicolumn{1}{c}{Method} & SAM     & Swin    & \multicolumn{2}{c}{ViT} & \multicolumn{2}{c}{MTAN}                     \\ \cmidrule(lr){2-2}  \cmidrule(lr){3-3} \cmidrule(lr){4-5} \cmidrule(lr){6-7}
                           & NYU-v2  & CelebA  & Clevr    & SmallNORB    & NYU-v2              & CityScapes             \\ 
\midrule
Joint Train w/ ST          & 30.97   & 37.60   & 29.38    & 29.38        & 62.19               &   76.02                     \\
PCGrad w/ ST               & 30.97   & 37.60   & 29.38    & 19.63        & 62.19               &    76.02                    \\
CAGrad w/ ST               & 30.97   & 72.85   & 29.38    & 29.38        & 62.19               &     76.02                   \\
GradDrop w/ ST             & 30.97   & 49.58   & 29.38    & 29.38        & 62.19               &    76.02                    \\
MGDA w/ ST                 & 30.97   & 37.60   & 29.38    & 29.38        & 62.19               &     62.19                   \\
IMTL-G w/ ST               & 30.97   & 37.60   & 29.38    & 29.38        & 62.19               &      62.19                  \\
NashMTL w/ ST              & 30.97   & 37.60   & 29.38    & 29.38        & 62.19               &    83.48                  \\
FAMO w/ ST                 & 30.97   & 37.60   & 29.38    & 29.38        & 62.19               &     62.19                   \\ 
\bottomrule
\end{tabular}
\caption{Number of trainable parameters. The values in the table are expressed as percentages (\%). As we select Top-K input parameters among all input connections for each neuron, therefore the same K might lead to different percentages of trainable parameters for different models. For example,  K=300 results in 30.97\% in SAM, 37.60\% in Swin, and 29.38\% in ViT for the pre-trained model.
}
\label{table: Number of trainable parameters}
\end{table*}

\subsection{Implementation details}
Following the work of Nash \cite{nash}, we apply all gradient manipulation techniques to the gradients of the shared weights. We set the hyperparameter c of CAGrad to 0.4, as it has been reported to yield optimal performance for NYUv2 and Cityscapes datasets \cite{CAGrad}. 
The experiments were conducted on the A100 80G GPU. Typically, training with SAM using NYU-v2 and Swin with CelebA requires approximately 1 day for a gradient manipulation method. Training ViT with SmallNORB takes around 18 minutes for a gradient manipulation method, while training ViT with Clevr takes about 30 minutes. On the other hand, training MTAN with NYU-v2 demands roughly 18 hours for a gradient manipulation method, whereas training MTAN with CityScapes necessitates approximately 12 hours.

\paragraph{SAM, ViT, Swin}
For all methods, including single-task learning, the gradient manipulation method, and our sparse training, we employed a batch size of 3 and searched for the optimal learning rate from the set \{2e-4, 5e-5\}, and then the best results are reported. The reason is that we find the optimal learning rate for sparse training is bigger than that for full parameters training. Therefore, for most methods, the optimal learning rate for sparse training is 2e-4 and that for the full parameters training is 5e-5. we also use data augmentations for all methods, following \cite{CAGrad}. The batch size used is set to be 3 for NYUv2 dataset, and 256 for CelebA, and 128 for SmallNORB and Clevr.

\paragraph{MTAN}
Following the works in \cite{nash, CAGrad}, we incorporate data augmentations during training for both \textit{Joint Train} method and all gradient manipulation methods. Each method is trained for 200 epochs with an initial learning rate of 0.0001, which is then reduced to 0.00005 after 100 epochs. For Multi-Task Learning (MTL) methods, we utilize a Multi-Task Attention Network (MTAN) \cite{SegNet_attention}  based on SegNet architecture proposed by \cite{badrinarayanan2017segnet}. Similar to previous studies \cite{CAGrad}, the STL baseline refers to training task-specific SegNet models. The batch size used is set to be 2 for NYUv2 dataset and 8 for CityScapes dataset respectively. To align with prior research on MTL including \cite{CAGrad, PCGrad, badrinarayanan2017segnet}, we report the test performance averaged over the last 10 epochs.

\section{Extended related work}
\label{sec: Extended related work}
\paragraph{Multi-task learning} 
Multi-task learning~\cite{zhang2021survey} aims to improve the overall performance of all tasks. 
In this work, we focus on a conventional setup of multi-task learning ~\cite{vandenhende2021multi}:  given a single input, multi-task models perform different and related predictions, such as segmentation, depth and surface normal. In other words, the input is shared by different tasks. 
In this paper, we roughly divide existing  MTL into two categories: 

\romannumeral1)
\textit{Multi-task optimization}. Recent works~\cite{PCGrad, CAGrad, Graddrop, MGDA, imtlg, nash, liu2023famo} provide impressive results in solving the task imbalance during optimization. The rationale behind these works is that re-weighting all task gradients or losses helps multi-task models reduce conflicting gradients among tasks~\cite{CAGrad, MGDA}. Specifically, some works~\cite{Graddrop, MGDA, imtlg} propose to form a new update gradient at each optimization by linearly combining task gradients. Other works~\cite{liu2023famo, kendall2018multi} learn dynamic loss scale to balance different tasks during training. However, it is challenging to scale up most existing optimization works to giant foundation models due to non-trivial computational and memory costs. In this paper, we propose a neuron-based parameter selection to sparsely fine-tune the pre-trained model, which boosts the performance of most optimization methods.

\romannumeral2)
\textit{Multi-task architecture}
In this branch, multi-task methods design different architectures to improve the exchanging or sharing of information among tasks~\cite{vandenhende2021multi}. Regarding where tasks interact, multi-task architectures are separated into \textit{encoder-focused} and \textit{decoder-focused}.
The former shares the information in the encoder by the transformation of activations among tasks~\cite{misra2016cross}, learnable task-specific attention modules~\cite{SegNet_attention}, branching networks for similar tasks~\cite{guo2020learning} and so on. 
The latter recursively uses task predictions to improve overall performance~\cite{xu2018pad, zhang2019pattern, zhang2018joint}. However, these architectures still suffer from the task imbalance issue during multi-task optimization. In this paper, our work focuses on boosting multi-task optimization.  As one of the multi-task optimization methods, our method can seamlessly generalize to different backbone models.

\paragraph{Training with subset of parameters}
several methods already proposed in single-task learning.
several methods select a subset of parameters based on a certain pre-defined rule, such as gradient \cite{zhang2023gradient, fu2023effectiveness} and magnitude of parameters \cite{lagunas2021block}. In addition to selecting parameters by hand design, \cite{sanh2020movement, mostafa2019parameter, xu2021raise} automatically select the subset of parameters through optimization. Although sparse training has been extensively investigated in single-task learning, its application in multi-task learning remains relatively unexplored. \cite{sun2020learning, calandriello2014sparse} learning to share information between tasks using a sparse model. Differing from them, in this paper, we systematically research the gradient conflict via the sparse training perspective.

\begin{table*}[t]
\centering
\resizebox{1\textwidth}{!}{
\begin{tabular}{cccccccccccccccccccc}
\toprule 
 & \multirow{3}{*}{\textbf{Methods}} & \multicolumn{2}{c}{Segmentation} & \multicolumn{2}{c}{Depth}  & \multicolumn{5}{c}{Surface Normal}  &  \multirow{3}{*}{$\mathbf{\Delta m \%} \downarrow$} \\
 \cmidrule(lr){3-4} \cmidrule(lr){5-6} \cmidrule(lr){7-11}
 &  &  \multirow{2}{*}{mIoU $\uparrow$} & \multirow{2}{*}{Pix Acc $\uparrow$}  & \multirow{2}{*}{Abs Err $\downarrow$} & \multirow{2}{*}{Rel Err $\downarrow$}  & \multicolumn{2}{c}{Angle Distance $\downarrow$} & \multicolumn{3}{c}{Within $t^\circ$  $\uparrow$} & &   \\
 \cmidrule(lr){7-8} \cmidrule(lr){9-11}
 &  &  &  &  &  &  Mean & Median  & 11.25  & 22.5  & 30 &  &  \\
\toprule
&\multicolumn{1}{c}{Random} & $59.85$& $80.09$& $0.3357$&$0.1359$ & $22.17$& $16.12$&  $36.08$&$64.50$ &$75.56$  &$6.014$ \\
& \multicolumn{1}{c}{Global} & $59.53$& $79.38$& $0.3380$&$0.1373$ & $22.32$& $16.32$ & $35.62$&$63.93$ &$75.16$ &$6.855$ \\
& \multicolumn{1}{c}{Reverse} & $59.35$& $79.57$& $0.3417$&$0.1396$ & $22.31$& $16.25$& $35.98$&$64.07$ &$75.16$ &$6.960$ \\
& \multicolumn{1}{c}{Ours} & $60.03$& $79.96$& $0.3320$&$0.1353$ & $21.98$& $15.92$& $36.69$&$64.92$ &$75.82$ &$5.314$ \\
\bottomrule
\end{tabular}%
}
\caption{Different sparse training methods on SAM model with NYU-v2 datasets.}
\label{tab:ablation (full): different sparse training methods}
\end{table*}

\begin{figure*}[t]
\centering
\subfloat[Ours]{
    \includegraphics[width=0.31\linewidth]{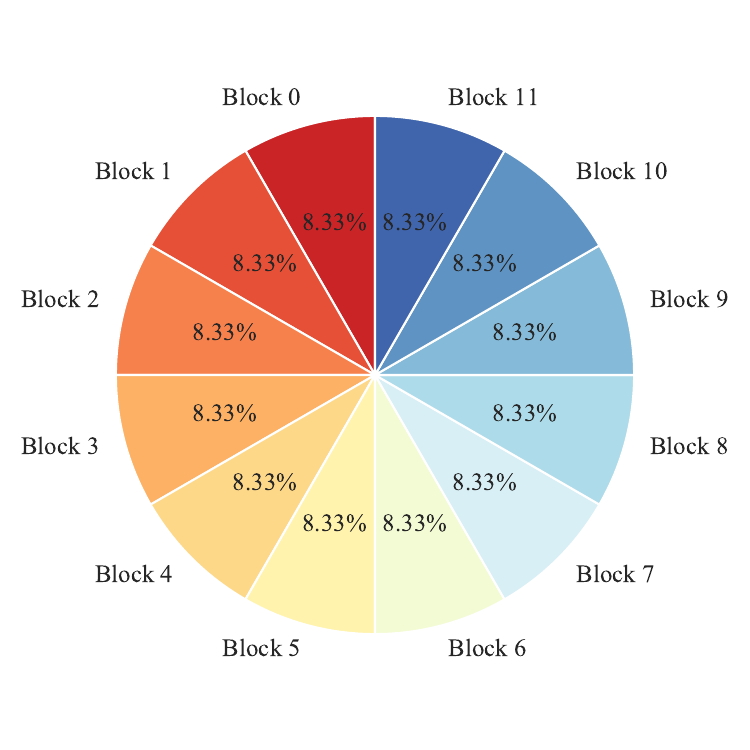}
    \label{fig: statics number of parameters for ours}
}
\hfill
\centering
\subfloat[Random]{
    \includegraphics[width=0.31\linewidth]{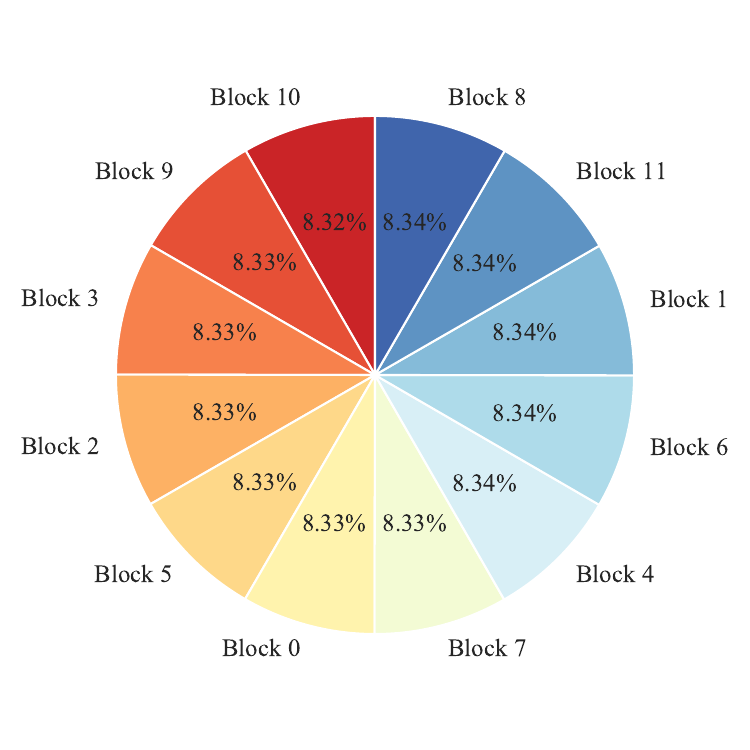}
    \label{fig: statics number of parameters for Random}
}
\hfill
\centering
\subfloat[Global]{
    \includegraphics[width=0.31\linewidth]{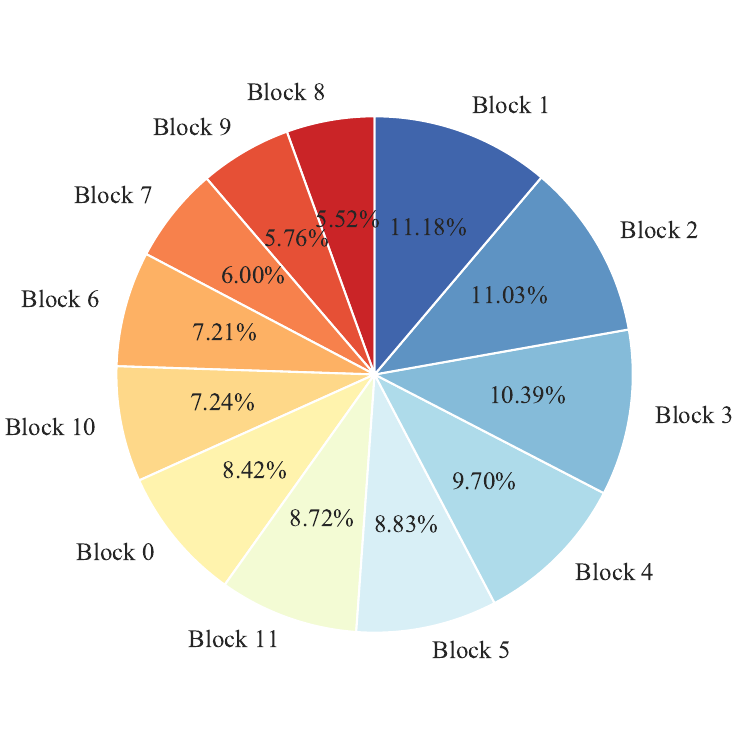}
    \label{fig: statics number of parameters for Global}
}
\caption{The distribution of selected trainable parameters for different sparse training methods over different blocks. The experiments are conducted on SAM model with NYU-v2 dataset.}
\label{fig: selected number of parameters for diff sparse methods}
\end{figure*}

\section{Detailed experiment results}
\label{sec: Detailed experiment results}
In this section, we provide the detailed experiment results conducted in the main body of our paper, including the average incident of gradient conflict, the incident of gradient conflict for all epochs, and visualization of the gradient conflict for \textit{Joint Train} and all gradient manipulation methods.

\subsection{Ablation study}
\label{sec: supp: Ablation study}
The detailed results for various sparse methods are provided in \cref{tab:ablation (full): different sparse training methods}, which is the full version of \cref{fig: different sparse method starplot}. It can be observed that, with the exception of \textit{Pix Acc} in segmentation, our sparse method outperforms other methods. In addition, we provide the distribution of the selected parameters using different sparse training over different blocks of the model. As shown in \cref{fig: selected number of parameters for diff sparse methods}, the parameters selected by our sparse training method and \textit{Random} are evenly distributed over the whole network. As for \textit{Global} selecting the parameters with the highest magnitude, the distribution of selected parameters is largely different over different blocks



\begin{figure*}[tb]
\centering
\subfloat[Joint Train]{
    \includegraphics[width=0.25\linewidth]{images/joint_gray.pdf}
    \label{fig:joint}
}
\subfloat[GradDrop]{
    \includegraphics[width=0.25\linewidth]{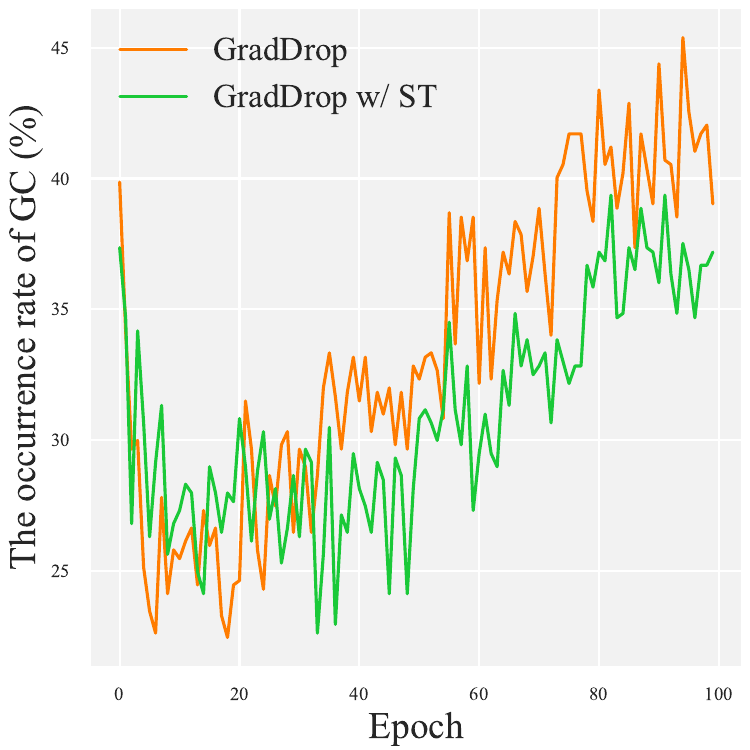}
    \label{fig:GradDrop}
}
\subfloat[IMTL-G]{
    \includegraphics[width=0.25\linewidth]{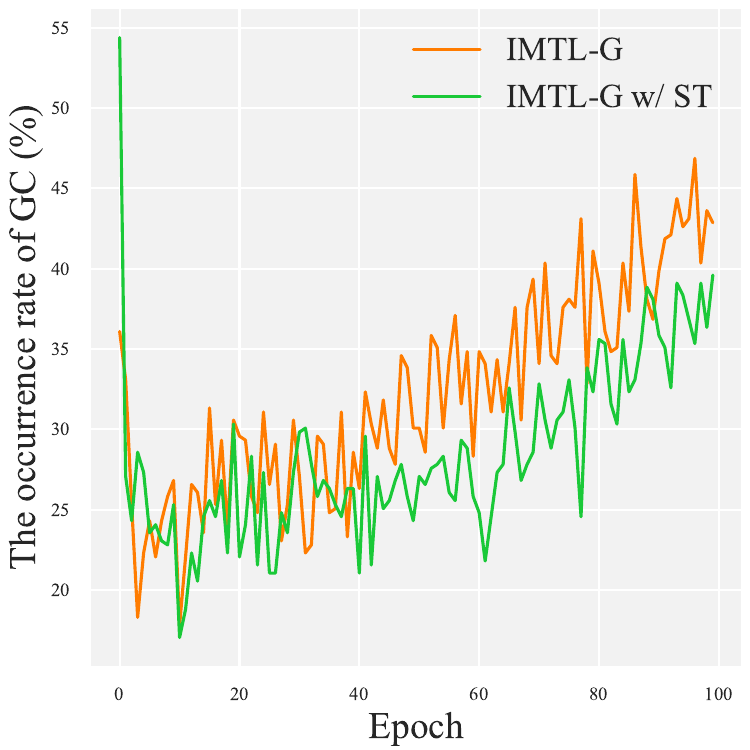}
    \label{fig:IMTLG}
} \\
\subfloat[CAGrad]{
    \includegraphics[width=0.25\linewidth]{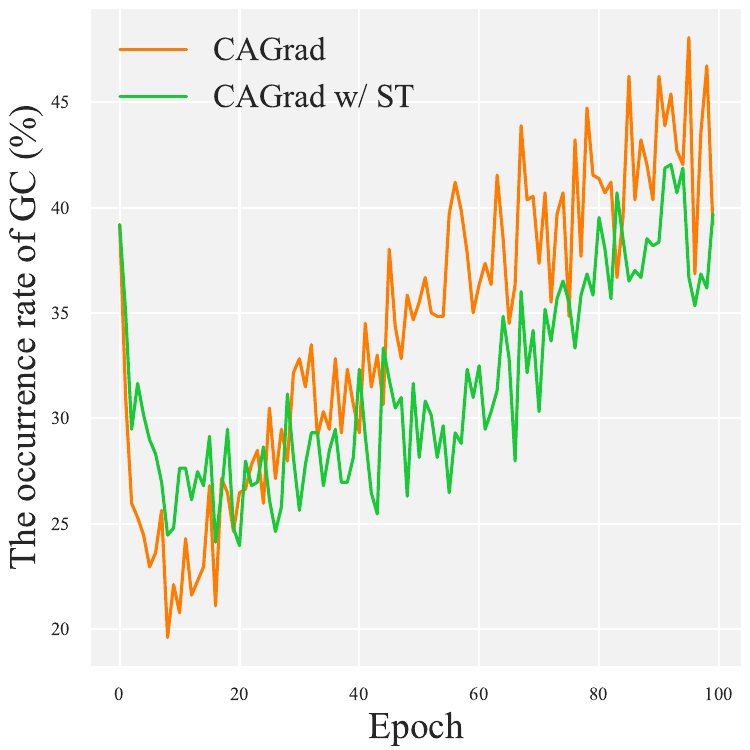}
    \label{fig:CAGrad}
}
\subfloat[PCGrad]{
    \includegraphics[width=0.25\linewidth]{images/pcgrad_gray.pdf}
    \label{fig:PCGrad}
}
\subfloat[NashMTL]{
    \includegraphics[width=0.25\linewidth]{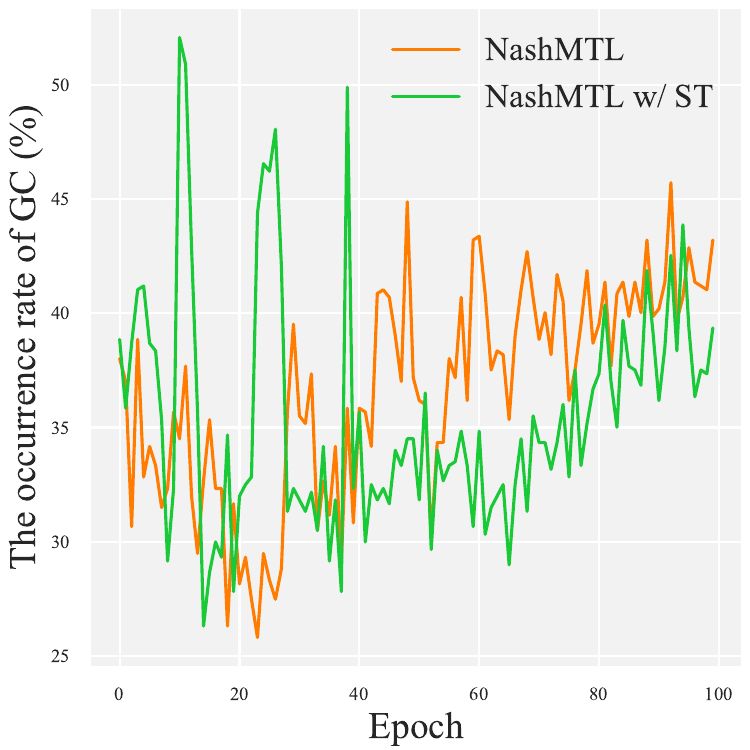}
    \label{fig:NashMTL}
}\\
\subfloat[MGDA]{
    \includegraphics[width=0.25\linewidth]{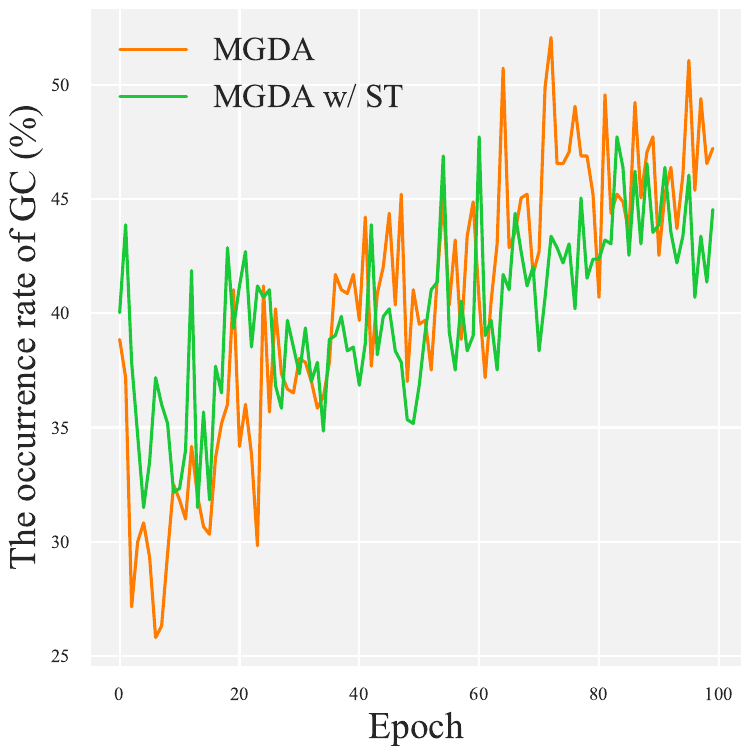}
    \label{fig:MGDA}
}
\caption{The number of occurrence gradient conflictions between tasks during training SAM on NYUv2 dataset.}
\label{fig:number of gc}
\end{figure*}

\begin{figure*}[tb]
\centering
\includegraphics[width=0.9\linewidth]{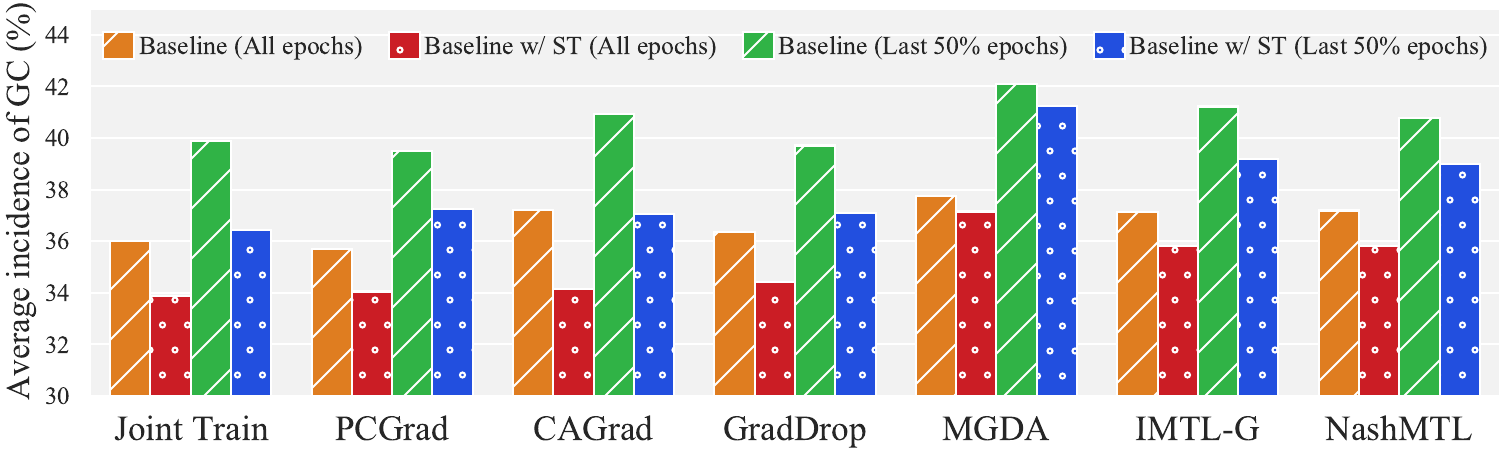}
\caption{
The average occurrence percentage of gradient conflict over epochs (all epochs/last 50\% epochs) during training on MTAN model with NYU-v2 datasets was evaluated using various methods, including joint training and gradient manipulation techniques.
}
\label{fig:gc of different methods mtan} 
\end{figure*}

\subsection{NYU-v2 on SAM}
\label{sec: supp NYU-v2 on SAM}
The incidence of gradient conflict for \textit{Joint Train} and gradient manipulation method  over all epochs are shown in \cref{fig:number of gc}, which is the full version of \cref{fig:incidence of gc (part)} in the main body of the paper.

\subsection{NYU-v2 on MTAN}
\label{sec: supp NYU-v2 on MTAN}

\begin{table*}[t]
        \centering
        \resizebox{0.4\textwidth}{!}{
        \begin{tabular}{ccll}
        \toprule 
         & \multirow{2}{*}{\textbf{Methods}}& \multicolumn{2}{c}{Average incidence of GC ($\%$)}\\
         \cmidrule(lr){3-4} &                                  &All epochs& Last 50\% epochs\\
        \toprule
        &Joint Train& 36.01&39.87 \\
        &     w/ ST & 33.86 (\textcolor{blue}{2.15}) &36.45 (\textcolor{blue}{3.42}) \\
        \midrule
         & PCGrad & 35.71& 39.51\\
         & w/ ST & 34.05 (\textcolor{blue}{1.66})&37.25 (\textcolor{blue}{2.26})\\
        \midrule
         & CAGrad& 37.21& 40.93\\
         & w/ ST& 34.14 (\textcolor{blue}{3.07})& 37.04 (\textcolor{blue}{3.89}) \\
        \midrule
         & GradDrop &36.37 & 39.71 \\
         & w/ ST & 34.42 (\textcolor{blue}{1.95})& 37.10(\textcolor{blue}{2.61})\\
        \midrule
         & MGDA & 37.76& 42.1\\
         & w/ ST& 37.15 (\textcolor{blue}{ 0.61})&41.25 (\textcolor{blue}{0.85})\\
        \midrule
         & IMTL-G &  37.14&41.22 \\
         & w/ ST & 35.81 (\textcolor{blue}{1.33})&39.17 (\textcolor{blue}{2.05})\\
        \midrule
         & NashMTL & 37.19& 40.79\\
         & w/ ST & 35.83 (\textcolor{blue}{1.36})& 39.0 (\textcolor{blue}{1.79})\\
        \bottomrule
        \end{tabular}%
        }
         \caption{Average incidence of gradient conflict between tasks over epochs for different methods. The improvement by sparse training is provided in (\textcolor{blue}{$\bullet$}). We calculate the average incidence of gradient conflict over all epochs and the last 50\% epochs during training MTAN on NYUv2.} 
        \label{tab:gc of different methods MTAN} 
\end{table*}

\begin{figure*}[t]
\subfloat[Joint Train]{
    \includegraphics[width=0.3\linewidth]{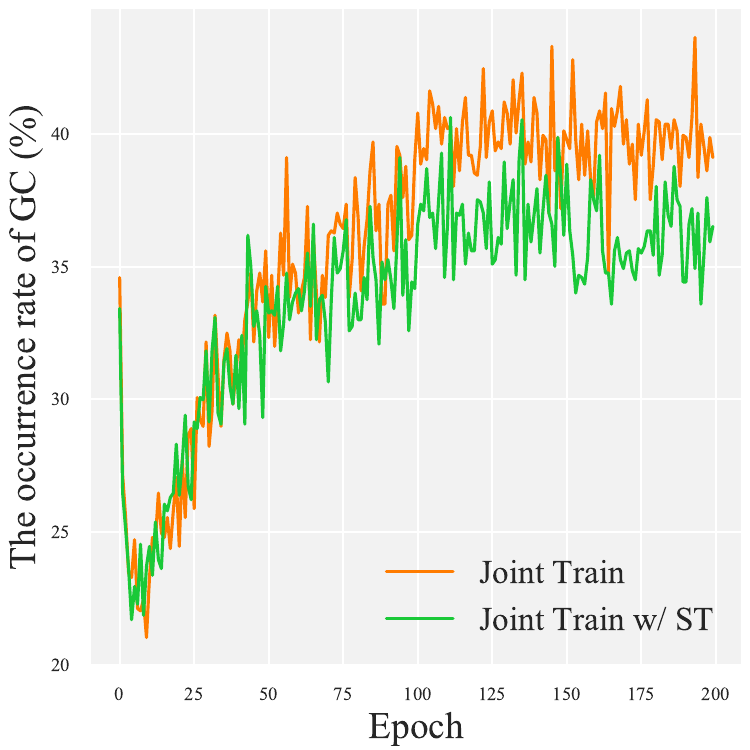}
    \label{fig:joint MTAN}
}
\subfloat[GradDrop]{
    \includegraphics[width=0.3\linewidth]{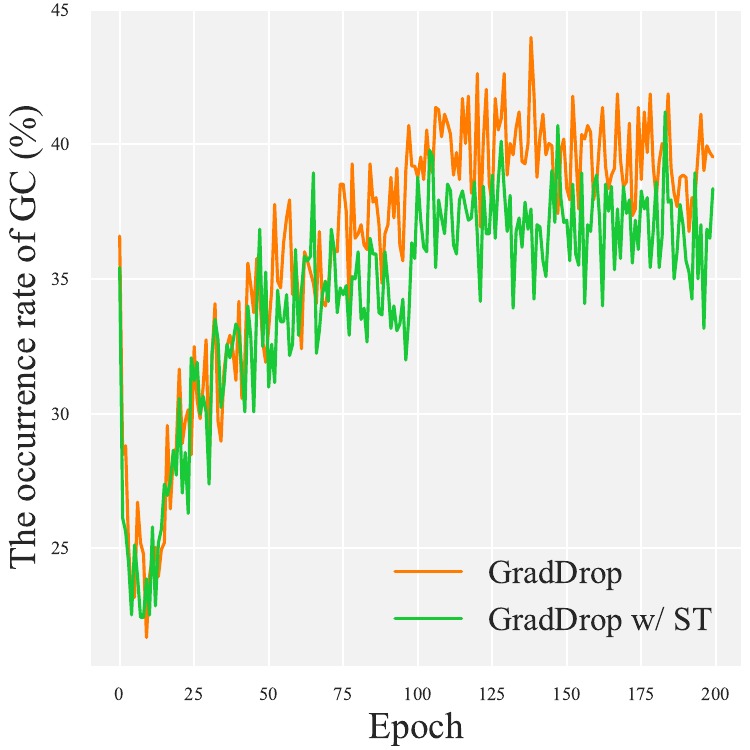}
    \label{fig:GradDrop MTAN}
}
\subfloat[IMTL-G]{
    \includegraphics[width=0.3\linewidth]{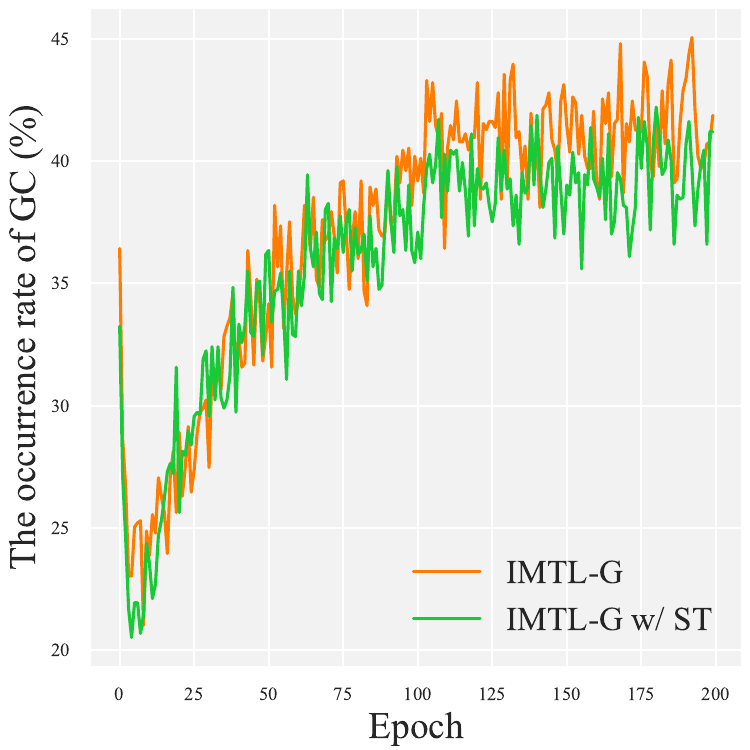}
    \label{fig:IMTLG MTAN}
} \\
\subfloat[CAGrad]{
    \includegraphics[width=0.3\linewidth]{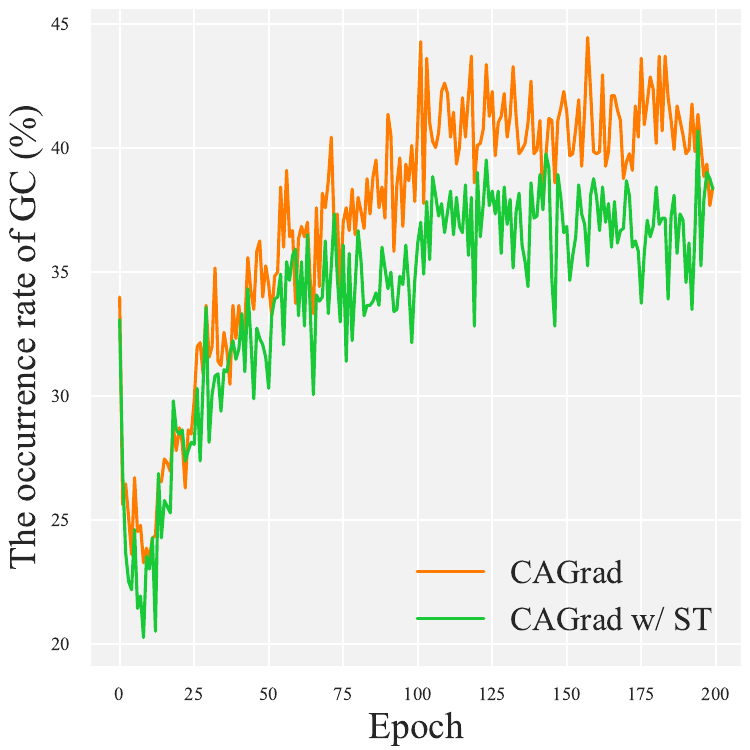}
    \label{fig:CAGrad MTAN}
}
\subfloat[PCGrad]{
    \includegraphics[width=0.3\linewidth]{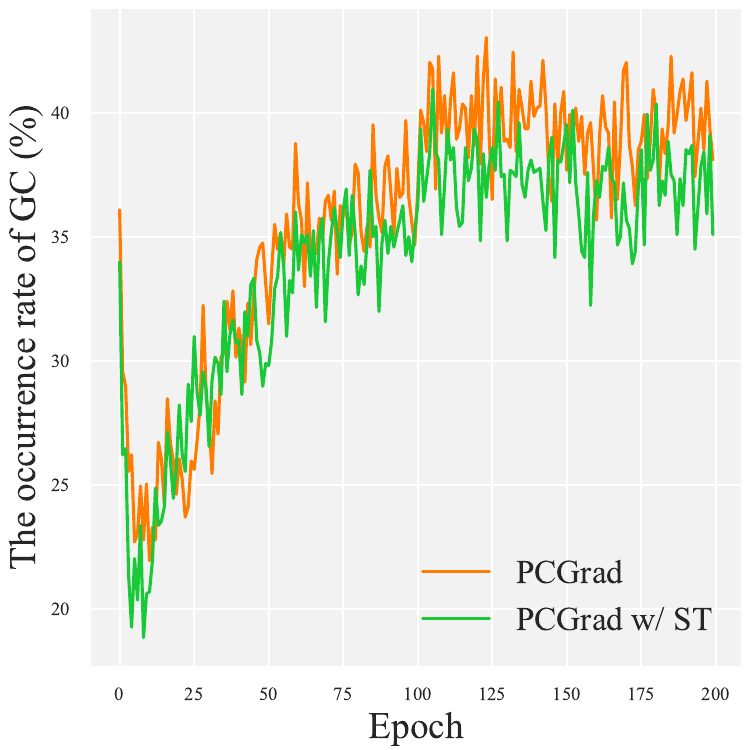}
    \label{fig:PCGrad MTAN}
}
\subfloat[NashMTL]{
    \includegraphics[width=0.3\linewidth]{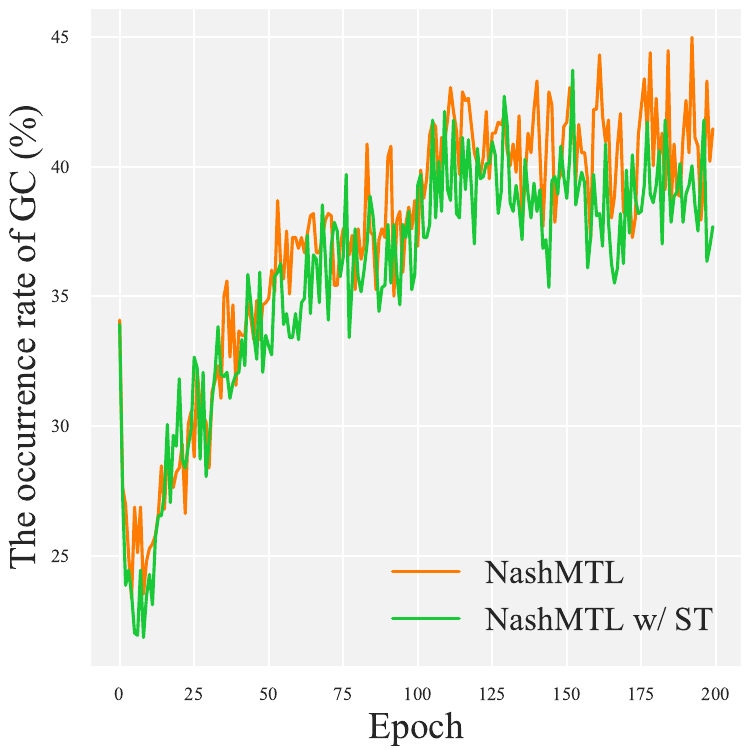}
    \label{fig:NashMTL MTAN}
}\\
\subfloat[MGDA]{
    \includegraphics[width=0.3\linewidth]{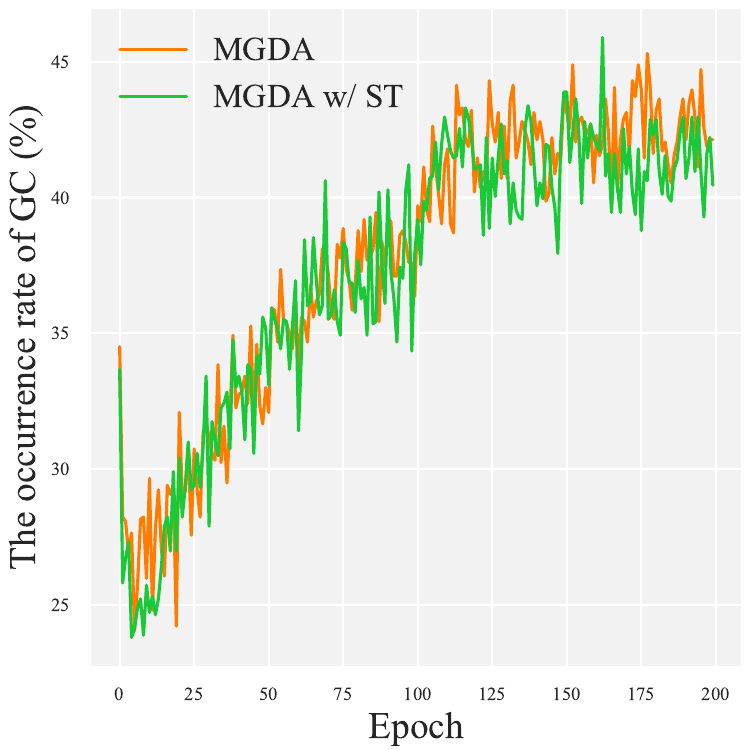}
    \label{fig:MGDA MTAN}
}
\caption{The number of occurrence gradient conflictions between tasks during tuning MTAN on NYUv2 dataset.}
\label{fig:number of gc mtan}
\end{figure*}

We also conduct experiments on MTAN with NYU-v2 dataset. MTAN is a random initialized model. As we can see in \cref{tab:gc of different methods MTAN}, even for the random initialized model, sparse training can also reduce the incidence of gradient conflict. The visualization of the occurrence of gradient conflict for each epoch is shown in 
\cref{fig:number of gc mtan} and the average incidence of gradient conflict across all epochs for different methods is shown in \cref{fig:gc of different methods mtan}. As for the performance of the overall tasks on NYU-v2, the sparse training improves not only the overall performance ($\Delta m\%$) but also the performance of each task for all methods including \textit{Joint Train} and all gradient manipulation methods, as shown in \cref{tab:nyu mtan}. In addition, following \cite{nash}, we conduct the experiments three times with three different seeds. The $mean$ ± $std$ is presented in \cref{tab:nyu mtan}, we can observe that the sparse training is robust to the random seed.

\begin{table*}[t]
    \centering
    \resizebox{\textwidth}{!}{%
    \begin{tabular}{cclllllllllcccc}
    \toprule 
     \multirow{3}{*}{\textbf{Methods}}   & & \multicolumn{2}{c}{Segmentation} & \multicolumn{2}{c}{Depth}  & \multicolumn{5}{c}{Surface Normal} &  \multirow{3}{*}{$\mathbf{\Delta m \%} \downarrow$} \\
     \cmidrule(lr){3-4} \cmidrule(lr){5-6} \cmidrule(lr){7-11}
      &  & \multirow{2}{*}{mIoU $\uparrow$} & \multirow{2}{*}{Pix Acc $\uparrow$}  & \multirow{2}{*}{Abs Err $\downarrow$} & \multirow{2}{*}{Rel Err $\downarrow$}   & \multicolumn{2}{c}{Angle Distance $\downarrow$} & \multicolumn{3}{c}{Within $t^\circ$  $\uparrow$} \\
     \cmidrule(lr){7-8} \cmidrule(lr){9-11}
     &  &  &  &  &  & Mean & Median  & 11.25 & 22.5  & 30 \\
    \toprule
    \multicolumn{2}{c}{STL}& 38.30 & 63.76 & 0.6754 & 0.2780 & 25.01 & 19.21 &  30.14 & 57.20 & 69.15   &$-$     \\
    \midrule
    \multicolumn{2}{c}{Joint Train} & 39.29 & 65.33 & 0.5493 & 0.2263 & 28.15 & 23.96 & 22.09 & 47.50 & 61.08   & 5.59 \\
    \multicolumn{2}{c}{w/ ST} &\cellcolor[HTML]{CCFFCC}41.04 (±0.28)&\cellcolor[HTML]{CCFFCC}66.05 (±0.12)&\cellcolor[HTML]{CCFFCC}0.5417 (±0.0008)&\cellcolor[HTML]{CCFFCC}0.2232 (±0.0011) &\cellcolor[HTML]{CCFFCC}27.40(±0.05) &\cellcolor[HTML]{CCFFCC}22.90(±0.12) &\cellcolor[HTML]{CCFFCC}23.58(±0.13)
    &\cellcolor[HTML]{CCFFCC}49.59(±0.14)&\cellcolor[HTML]{CCFFCC}63.01(±0.09)&\cellcolor[HTML]{CCFFCC}2.49(±0.11) \\
    \midrule
    \multicolumn{2}{c}{PCGrad}     & 38.06 & 64.64 &0.5550 & 0.2325 & 27.41 & 22.80& 23.86 &49.83 & 63.14 & 3.97  \\
    \multicolumn{2}{c}{w/ ST} &\cellcolor[HTML]{CCFFCC}40.49 (±0.32)&\cellcolor[HTML]{CCFFCC}66.17(±0.23)  &\cellcolor[HTML]{CCFFCC}0.5441 (±0.0023)&\cellcolor[HTML]{CCFFCC}0.2264  (±0.0030)&\cellcolor[HTML]{CCFFCC}27.09 (±0.08)& \cellcolor[HTML]{CCFFCC}22.55(±0.03)&\cellcolor[HTML]{CCFFCC}24.22(±0.12) &\cellcolor[HTML]{CCFFCC}50.34(±0.17)&\cellcolor[HTML]{CCFFCC}63.63(±0.12)&\cellcolor[HTML]{CCFFCC}1.98(±0.12) \\
    \midrule
    \multicolumn{2}{c}{CAGrad}    & 39.79 & 65.49 & 0.5486 & 0.2250 & 26.31 & 21.58 & 25.61 & 52.36 & 65.58   & 0.20  \\
    \multicolumn{2}{c}{w/ ST} & \cellcolor[HTML]{CCFFCC}39.93(±0.33)&\cellcolor[HTML]{CCFFCC}66.19(±0.16) &\cellcolor[HTML]{CCFFCC}0.5299(±0.0025) &\cellcolor[HTML]{CCFFCC}0.2097(±0.0038) &\cellcolor[HTML]{CCFFCC}25.71(±0.02) &\cellcolor[HTML]{CCFFCC}20.70(±0.03) &\cellcolor[HTML]{CCFFCC}26.86(±0.13) &\cellcolor[HTML]{CCFFCC}54.22(±0.15) &\cellcolor[HTML]{CCFFCC}67.30(±0.13) &\cellcolor[HTML]{CCFFCC}-2.76(±0.10) \\
    \midrule
    \multicolumn{2}{c}{GradDrop}  & 39.39 & 65.12 & 0.5455 & 0.2279 & 27.48 & 22.96 & 23.38 & 49.44 & 62.87 & 3.58  \\
    \multicolumn{2}{c}{w/ ST} &\cellcolor[HTML]{CCFFCC}40.84(±0.35) &\cellcolor[HTML]{CCFFCC}66.84(±0.24) &\cellcolor[HTML]{CCFFCC}0.5288(±0.0021) &\cellcolor[HTML]{CCFFCC}0.2209(±0.0021) &\cellcolor[HTML]{CCFFCC}27.18(±0.03) &\cellcolor[HTML]{CCFFCC}22.56(±0.07) &\cellcolor[HTML]{CCFFCC}24.10(±0.11) &\cellcolor[HTML]{CCFFCC}50.33(±0.14) &\cellcolor[HTML]{CCFFCC}63.67(±0.13) &\cellcolor[HTML]{CCFFCC}1.38(±0.12) \\
    \midrule
    \multicolumn{2}{c}{MGDA}      & 30.47 & 59.90 & 0.6070 & 0.2555 & 24.88 & 19.45 & 29.18& 56.88 & 69.36  & 1.38  \\
    \multicolumn{2}{c}{w/ ST} &\cellcolor[HTML]{CCFFCC}32.42(±0.41) & \cellcolor[HTML]{CCFFCC}61.61(±0.21)&\cellcolor[HTML]{CCFFCC}0.5851(±0.0015) &\cellcolor[HTML]{CCFFCC}0.2239 (±0.0032)&\cellcolor[HTML]{CCFFCC}24.35(±0.02) &\cellcolor[HTML]{CCFFCC}18.61(±0.03) &\cellcolor[HTML]{CCFFCC}31.14(±0.12) &\cellcolor[HTML]{CCFFCC}58.63(±0.15) &\cellcolor[HTML]{CCFFCC}70.62(±0.13) &\cellcolor[HTML]{CCFFCC}-3.09(±0.14) \\
    \midrule
    \multicolumn{2}{c}{IMTL-G}     & 39.35 & 65.60 & 0.5426 & 0.2256 & 26.02 & 21.19 & 26.20 & 53.13 & 66.24   & -0.76  \\
    \multicolumn{2}{c}{w/ ST} &\cellcolor[HTML]{CCFFCC}40.73(±0.33) &\cellcolor[HTML]{CCFFCC}66.00(±0.17) &\cellcolor[HTML]{CCFFCC}0.5219(±0.0015) &\cellcolor[HTML]{CCFFCC}0.2100(±0.0021) &\cellcolor[HTML]{CCFFCC}25.6(±0.05) &\cellcolor[HTML]{CCFFCC}20.64(±0.04) &\cellcolor[HTML]{CCFFCC}26.81(±0.16) &\cellcolor[HTML]{CCFFCC}54.38(±0.15) &\cellcolor[HTML]{CCFFCC}67.49(±0.12) &\cellcolor[HTML]{CCFFCC}-3.18(±0.11) \\
    \midrule
    \multicolumn{2}{c}{NashMTL}   & 40.13 & 65.93 & 0.5261 & 0.2171 & 25.26 & 20.08 & 28.40 & 55.47 & 68.15 & -4.04  \\
    \multicolumn{2}{c}{w/ ST} & 39.75(±0.21)&\cellcolor[HTML]{CCFFCC}66.45(±0.05) &\cellcolor[HTML]{CCFFCC}0.5156(±0.0006)&\cellcolor[HTML]{CCFFCC}0.2121(±0.0009) &\cellcolor[HTML]{CCFFCC}24.96(±0.01) &\cellcolor[HTML]{CCFFCC}19.80(±0.05) &\cellcolor[HTML]{CCFFCC}28.80(±0.11) &\cellcolor[HTML]{CCFFCC}56.20(±0.10) &\cellcolor[HTML]{CCFFCC}68.93(±0.09) &\cellcolor[HTML]{CCFFCC}-5.11(±0.07) \\
    \bottomrule  
    \end{tabular}
    }
    \caption{The test performance on NYU-v2 dataset training on MTAN model, involving three tasks: semantic segmentation, depth estimation and surface normal. The result is the mean over three random seeds (\textit{std} is presented in  (± $\bullet$). The green cell color indicates that sparse training improves the performance of joint training or gradient manipulation methods. The best result is highlighted in bold.}
    \label{tab:nyu mtan} 
\end{table*}

\subsection{NYU-v2 on Swin}
\label{sec: supp NYU-v2 on Swin}

In order to investigate how the incidence of gradient conflict changes with varying model sizes, we conduct experiments on Swin/Tiny, Swin/Base and Swin/Large through the \textit{Joint Train}. As depicted in \cref{tab:gc of different swin model size}, there is an observed increase in the incidence of gradient conflict as the model size increases. Additionally, the performance of tasks improves as the model size increases \cref{tab:nyu swin performance}.

\begin{table}[t]
\centering
\begin{tabular}{cccc}
\toprule 
 & \multicolumn{2}{c}{\textbf{Model / Size}}& Average incidence of GC (\%)\\
\midrule
 & \multicolumn{2}{c}{Swin / Tiny }& 37.42 \\
 & \multicolumn{2}{c}{Swin / Base }& 40.34\\
 & \multicolumn{2}{c}{Swin / Large}& 41.84 \\
\bottomrule
\end{tabular}%
\caption{The average incidence of gradient conflict across all epochs during joint training with NYU-v2 on different sizes of Swin transformer.}
\label{tab:gc of different swin model size}
\end{table}

\begin{table*}[t]
    \centering
    \resizebox{\textwidth}{!}{%
    \begin{tabular}{cccccccccccccc}
    \toprule 
     \multirow{3}{*}{\textbf{Model}}   & & \multicolumn{2}{c}{Segmentation} & \multicolumn{2}{c}{Depth}  & \multicolumn{5}{c}{Surface Normal}\\
     \cmidrule(lr){3-4} \cmidrule(lr){5-6} \cmidrule(lr){7-11}
      &  & \multirow{2}{*}{mIoU $\uparrow$} & \multirow{2}{*}{Pix Acc $\uparrow$}  & \multirow{2}{*}{Abs Err $\downarrow$} & \multirow{2}{*}{Rel Err $\downarrow$}   & \multicolumn{2}{c}{Angle Distance $\downarrow$} & \multicolumn{3}{c}{Within $t^\circ$  $\uparrow$} \\
     \cmidrule(lr){7-8} \cmidrule(lr){9-11}
     &  &  &  &  &  & Mean & Median  & 11.25 & 22.5  & 30 \\
    \toprule
    \multicolumn{2}{c}{Swin/Tiny}&55.22&76.54&0.3746&0.1542&27.47&21.70&27.81&52.40&64.05  \\
    \multicolumn{2}{c}{Swin/Base}&59.60&79.16&0.3419&0.1388&25.88&19.74&31.23&56.24&67.32  \\
    \multicolumn{2}{c}{Swin/Large}& 61.34&80.28&0.3321&0.1345&25.09&18.73&33.05&58.12&68.86 \\
    \bottomrule  
    \end{tabular}
    }
    \caption{The test performance on NYU-v2 dataset jointly training on Swin models. }
    \label{tab:nyu swin performance} 
\end{table*}

\subsection{CelebA on Swin}
\label{sec: supp CelebA on Swin}

\begin{figure*}[t]
\centering
\subfloat[Joint Train]{
    \includegraphics[width=0.25\linewidth]{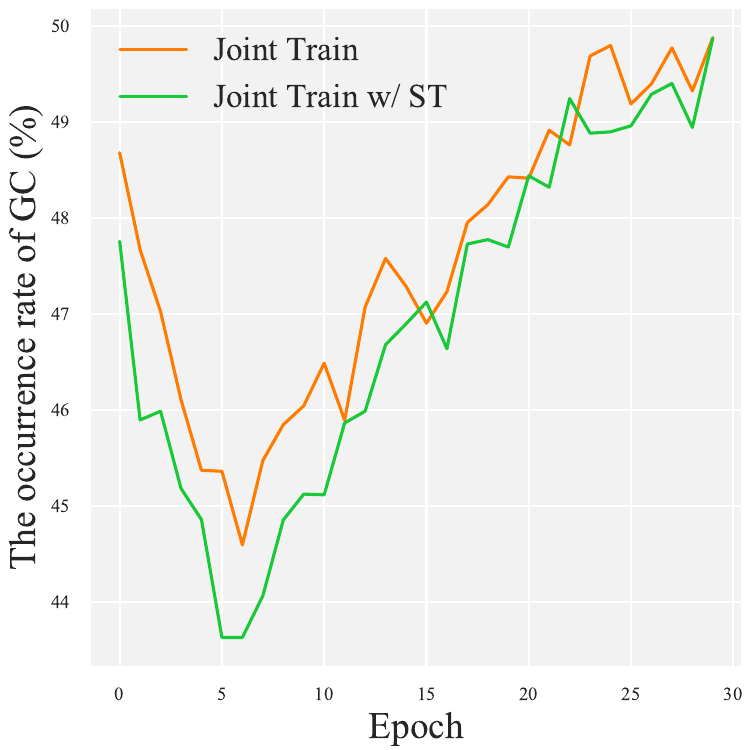}
    \label{fig:joint _celeba_swin}
}
\subfloat[GradDrop]{
    \includegraphics[width=0.25\linewidth]{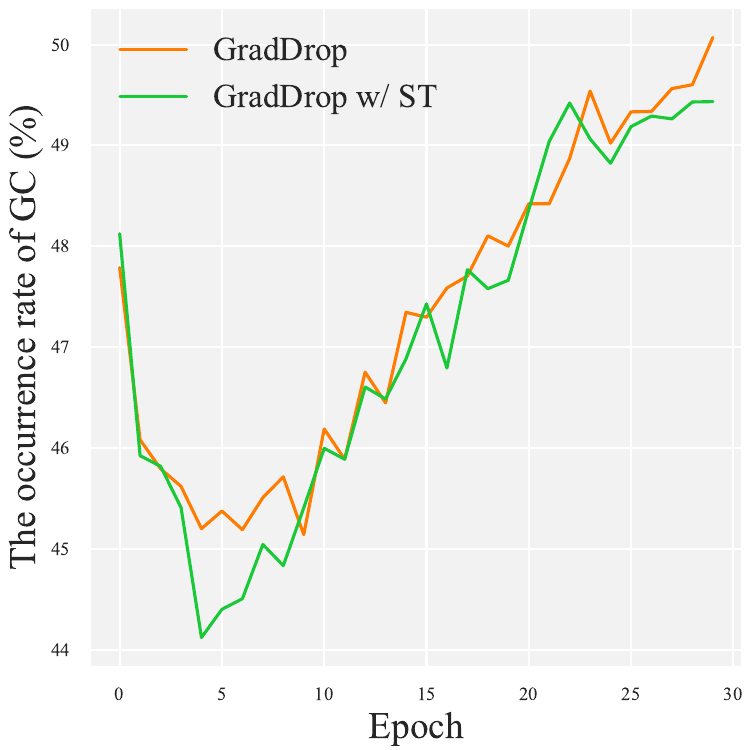}
    \label{fig:GradDrop _celeba_swin}
}
\subfloat[IMTL-G]{
    \includegraphics[width=0.25\linewidth]{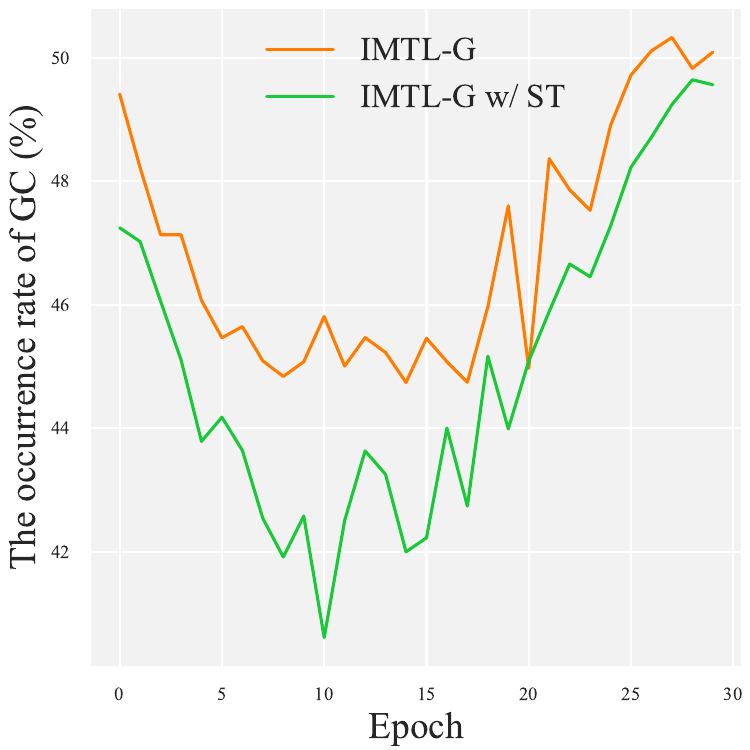}
    \label{fig:IMTLG _celeba_swin}
} \\
\subfloat[CAGrad]{
    \includegraphics[width=0.25\linewidth]{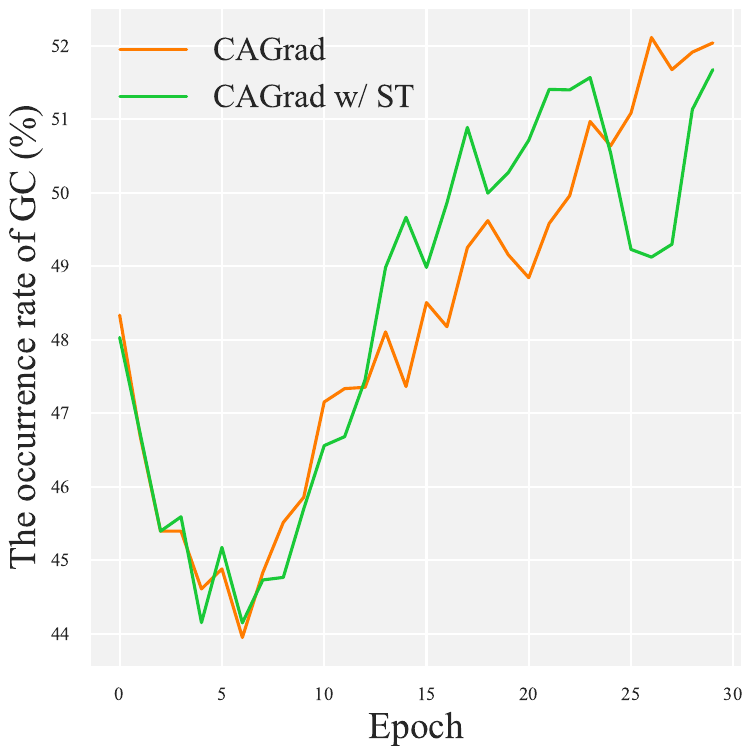}
    \label{fig:CAGrad _celeba_swin}
}
\subfloat[PCGrad]{
    \includegraphics[width=0.25\linewidth]{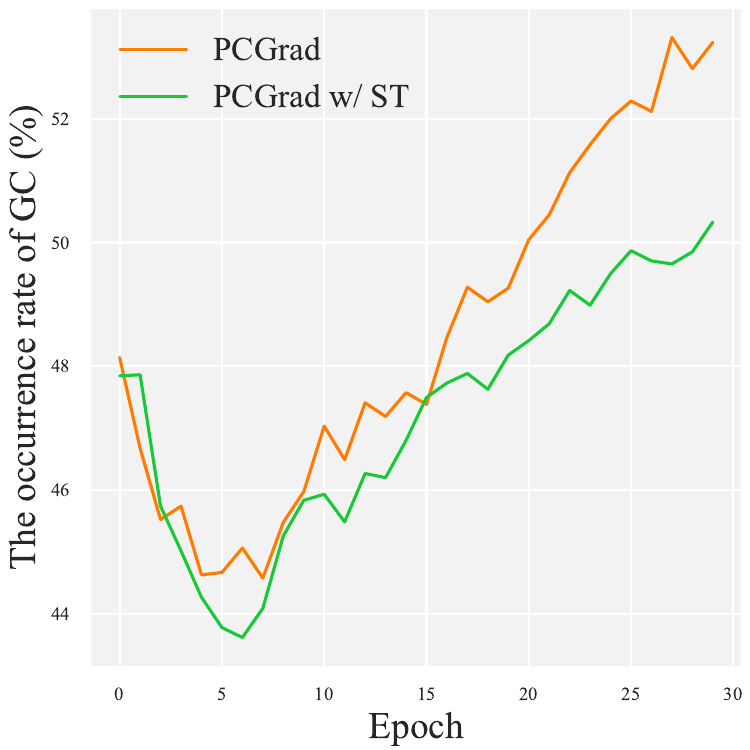}
    \label{fig:PCGrad _celeba_swin}
}
\subfloat[NashMTL]{
    \includegraphics[width=0.25\linewidth]{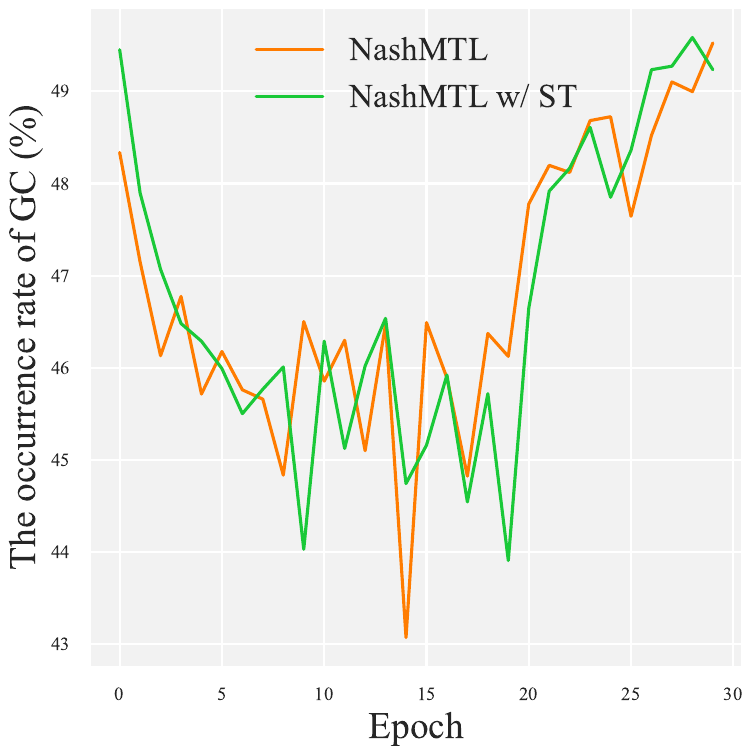}
    \label{fig:NashMTL _celeba_swin}
}\\
\subfloat[MGDA]{
    \includegraphics[width=0.25\linewidth]{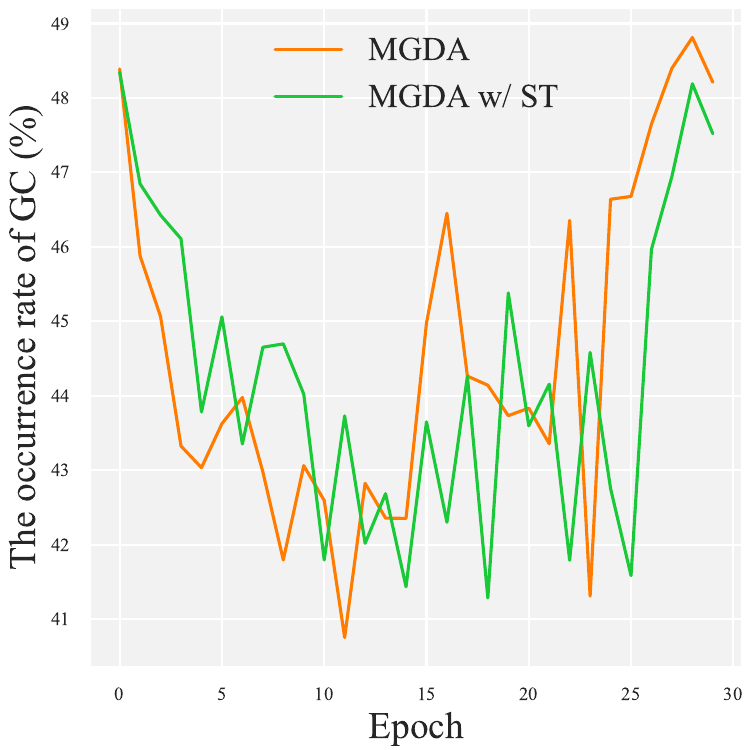}
    \label{fig:MGDA _celeba_swin}
}
\caption{The number of occurrence gradient conflictions between tasks during tuning Swin on CelebA dataset.}
\label{fig:number of gc Swin on CelebA} 
\end{figure*}

\begin{figure*}[t]
\centering
\includegraphics[width=1\linewidth]{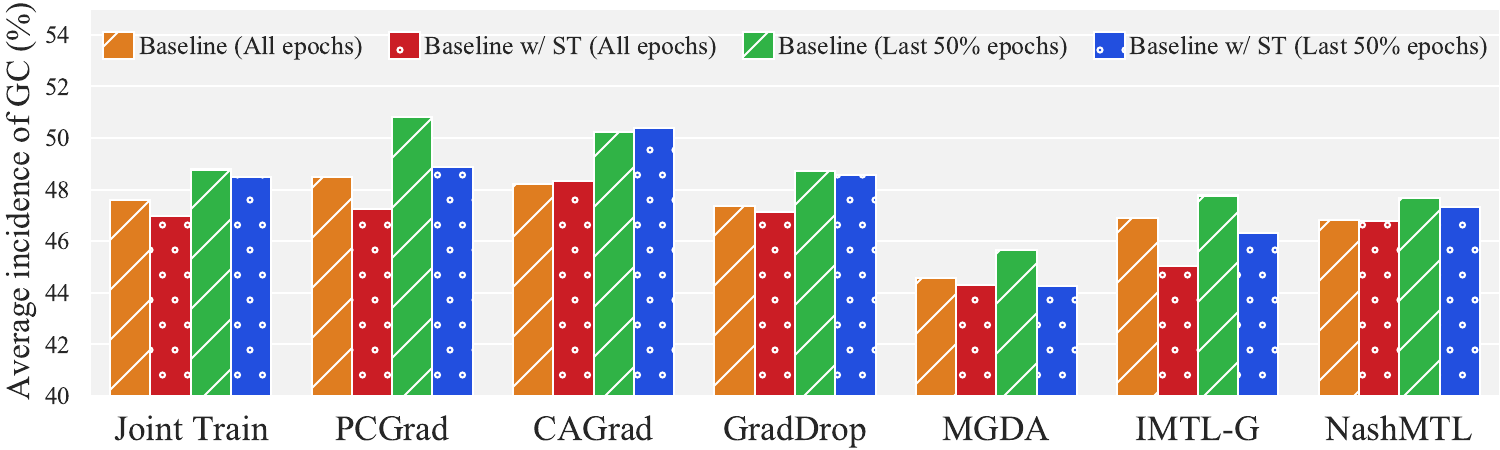}
\caption{
The average occurrence percentage of gradient conflict over epochs (all epochs/last 50\% epochs) during training on Swin model with CelebA datasets was evaluated using various methods, including joint training and gradient manipulation techniques.
}
\label{fig:gc of different methods CelebA on Swin} 
\end{figure*}

Following \cite{nash}, we train CelebA on Swin for only 30 epochs, because there are many more tasks in this dataset compared with other datasets, which leads to a significant increase in computation. As we can observe in \cref{tab:gc of different methods celeba swin}, most of the methods including \textit{Joint Train} and gradient manipulation methods can be improved by sparse training in terms of average incidence of gradient conflict between tasks over epochs.  It is noted that the improvement by sparse training here is not significant, which is because of the limited training epoch. Specifically, as shown in \cref{tab:gc of different methods} and \cref{tab:gc of different methods MTAN}, our sparse training improves more for later epochs. As for the performance of CelebA on Swin, please refer to \cref{tab:celeba clever SmallNORB nyuv2 city}. The visualization for the occurrence of gradient conflict for each epoch and average incidence of gradient conflict over all epochs for different methods, including \textit{Joint Train} and all gradient manipulation methods, are shown in \cref{fig:number of gc Swin on CelebA} and \cref{fig:gc of different methods CelebA on Swin}

\begin{table*}[t]
        \centering
        \resizebox{0.4\textwidth}{!}{
        \begin{tabular}{ccll}
        \toprule 
         & \multirow{2}{*}{\textbf{Methods}}& \multicolumn{2}{c}{Average incidence of GC ($\%$)}\\
         \cmidrule(lr){3-4} &                                  &All epochs& Last 50\% epochs\\
        \toprule
        &Joint Train& 47.61&48.78 \\
        &     w/ ST & 46.96 (\textcolor{blue}{0.65}) &48.48 (\textcolor{blue}{0.30}) \\
        \midrule
         & PCGrad & 48.48&50.83 \\
         & w/ ST & 47.24  (\textcolor{blue}{1.24}) &48.88 (\textcolor{blue}{1.95}) \\
        \midrule
         & CAGrad&48.21  & 50.23\\
         & w/ ST& 48.33  (\textcolor{blue}{-0.12}) & 50.40(\textcolor{blue}{-0.17}) \\
        \midrule
         & GradDrop & 47.36& 48.72\\
         & w/ ST &  47.13 (\textcolor{blue}{0.23}) &48.57 (\textcolor{blue}{0.15}) \\
        \midrule
         & MGDA & 44.56& 45.65 \\
         & w/ ST& 44.30  (\textcolor{blue}{0.26}) & 44.26(\textcolor{blue}{1.39}) \\
        \midrule
         & IMTL-G &46.89 & 47.77\\
         & w/ ST &  45.03 (\textcolor{blue}{1.86}) & 46.32(\textcolor{blue}{1.45}) \\
        \midrule
         & NashMTL &  46.83& 47.67\\
         & w/ ST &   46.78(\textcolor{blue}{0.05}) & 47.34(\textcolor{blue}{0.33}) \\
        \bottomrule
        \end{tabular}%
        }
        \caption{Average incidence of gradient conflict between tasks over epochs for different methods. The improvement by sparse training is provided in (\textcolor{blue}{$\bullet$}). We calculate the average incidence of gradient conflict over all epochs and the last 50\% epochs during training Swin on CelebA.} 
        \label{tab:gc of different methods celeba swin} 
\end{table*}

\subsection{SmallNORB on ViT}
\label{sec: supp SmallNORB on ViT}

SmallNORB is a much more difficult benchmark compared to other benchmarks in this paper. It comprises artificial objects observed under varying conditions and includes two tasks: object azimuth
and camera-elevation prediction. As shown in \cref{tab:smallnorb}, even for the \textit{STL}, the Top 1 accuracy only achieves $\sim$30\%, therefore, we use Top 5 as an extra metric here. We observed that even for this difficult task, sparse training can still achieve better performance compared with \textit{Joint Train} and all gradient manipulation methods.

\begin{table*}[t]
\centering
\begin{tabular}{ccccccccccccc}
\toprule 
 & \multirow{2}{*}{\textbf{Methods}}& &\multicolumn{2}{c}{Object Azimuth}   & \multicolumn{2}{c}{Camera Elevation} & \multirow{2}{*}{$\mathbf{\Delta m \%} \downarrow$}\\
 \cmidrule(lr){4-5} \cmidrule(lr){6-7} 
 &  &  &Top 1 $\uparrow$ &Top 5 $\uparrow$   &  Top 1 $\uparrow$ & Top 5 $\uparrow$ \\
\toprule
& \multicolumn{2}{c}{STL} &$32.92$ &$70.06$  &$36.56$& $94.67$& $-$&  \\
\midrule
& \multicolumn{2}{c}{Joint Train} &$28.01$ & $67.05$ & $29.84$& $89.75$&  $10.70$\\
& \multicolumn{2}{c}{w/ ST } & $27.33$& \cellcolor[HTML]{CCFFCC}$68.35$ & \cellcolor[HTML]{CCFFCC}$30.73$&\cellcolor[HTML]{CCFFCC}$89.87$ & \cellcolor[HTML]{CCFFCC}$10.11$  \\
\midrule
& \multicolumn{2}{c}{ PCGrad} & $28.79$& $67.85$ & $30.10$&$88.44$ & $9.99$  \\
& \multicolumn{2}{c}{w/ ST} &$27.539$ & \cellcolor[HTML]{CCFFCC}$67.92$ & \cellcolor[HTML]{CCFFCC}$31.18$& \cellcolor[HTML]{CCFFCC}$90.18$&  $\cellcolor[HTML]{CCFFCC}9.71$\\
\midrule
& \multicolumn{2}{c}{CAGrad } & $28.72$& $68.42$&  $29.33$&$87.93$ & $10.50$  \\
& \multicolumn{2}{c}{w/ ST} &$28.59$ & $68.21$&  \cellcolor[HTML]{CCFFCC}$29.82$& \cellcolor[HTML]{CCFFCC}$88.37$&  \cellcolor[HTML]{CCFFCC}$10.22$\\
\midrule
& \multicolumn{2}{c}{GradDrop } & $27.50$& $66.13$&  $29.86$&$88.50$ & $11.73$  \\
& \multicolumn{2}{c}{w/ ST} &\cellcolor[HTML]{CCFFCC}$28.34$ & \cellcolor[HTML]{CCFFCC}$67.79$ & $29.52$& $88.38$&  \cellcolor[HTML]{CCFFCC}$10.76$\\
\midrule
& \multicolumn{2}{c}{MGDA} & $30.82$& $70.13$&  $27.29$&$86.16$ & $10.15$  \\
& \multicolumn{2}{c}{w/ ST} &$28.28$ & $68.88$&  \cellcolor[HTML]{CCFFCC}$30.01$& \cellcolor[HTML]{CCFFCC}$89.47$&  \cellcolor[HTML]{CCFFCC}$9.79$\\
\midrule
& \multicolumn{2}{c}{IMTL-G } & $29.57$& $69.92$ & $28.51$&$86.74$ & $10.19$  \\
& \multicolumn{2}{c}{w/ ST} &$27.65$ & $69.09$&  \cellcolor[HTML]{CCFFCC}$30.01$& \cellcolor[HTML]{CCFFCC}$89.66$&  \cellcolor[HTML]{CCFFCC}$10.15$\\
\midrule
& \multicolumn{2}{c}{NashMTL } & $27.02$& $66.88$&  $30.83$&$89.74$ & $10.84$  \\
& \multicolumn{2}{c}{w/ ST} &\cellcolor[HTML]{CCFFCC}$28.17$ & \cellcolor[HTML]{CCFFCC}$67.93$& \cellcolor[HTML]{CCFFCC}$31.01$& $89.35$&  \cellcolor[HTML]{CCFFCC}$9.57$\\
\bottomrule
\end{tabular}%
\caption{The test performance on SmallNORB dataset trained on ViT. The green cell color indicates that sparse training improves the performance of joint training or gradient manipulation methods. The best result is highlighted in bold. }
\label{tab:smallnorb}
\end{table*}

\subsection{CityScapes on MTAN}
\label{sec: supp CityScapes on MTAN}

\begin{figure*}[t]
\centering
\subfloat[Joint Train]{
    \includegraphics[width=0.25\linewidth]{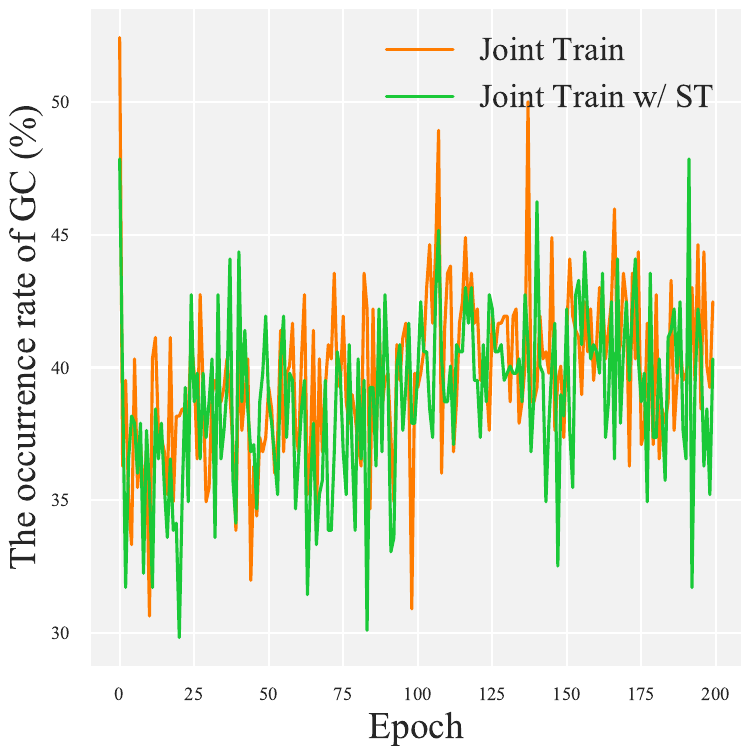}
    \label{fig:joint joint_gray_mtan_city}
}
\subfloat[GradDrop]{
    \includegraphics[width=0.25\linewidth]{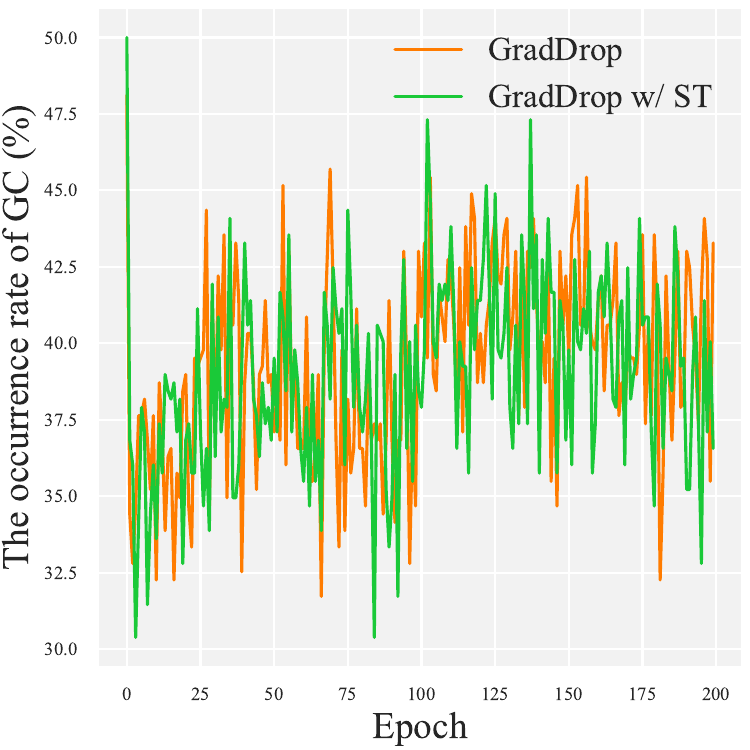}
    \label{fig:graddrop_gray_mtan_city}
}
\subfloat[IMTL-G]{
    \includegraphics[width=0.25\linewidth]{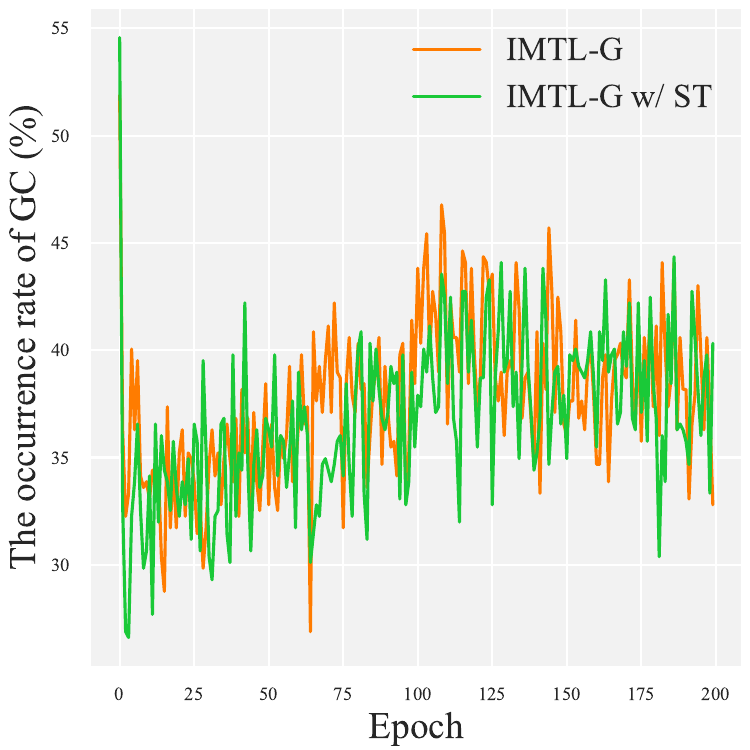}
    \label{fig:IMTLG imtlg_gray_mtan_city}
} \\
\subfloat[CAGrad]{
    \includegraphics[width=0.25\linewidth]{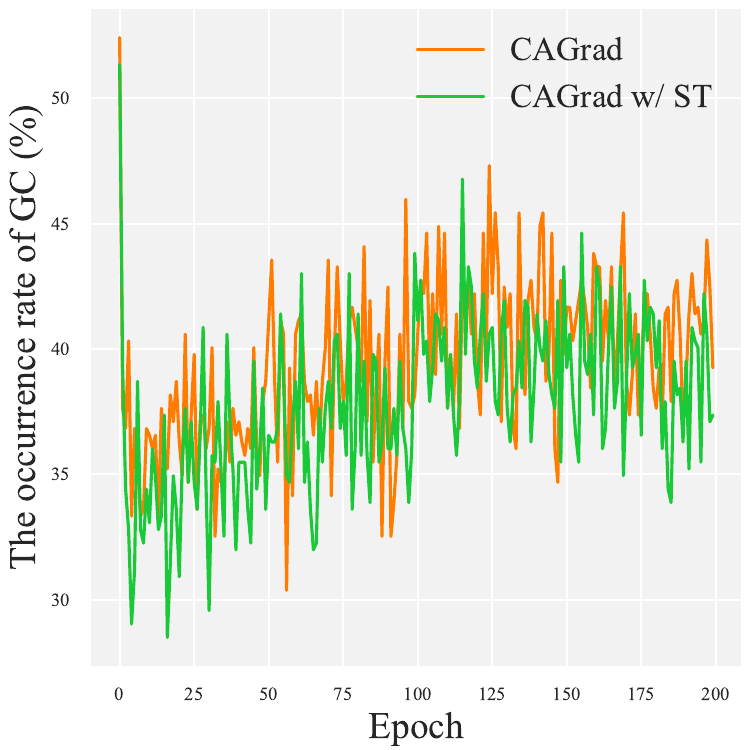}
    \label{fig:CAGrad cagrad_gray_mtan_city}
}
\subfloat[PCGrad]{
    \includegraphics[width=0.25\linewidth]{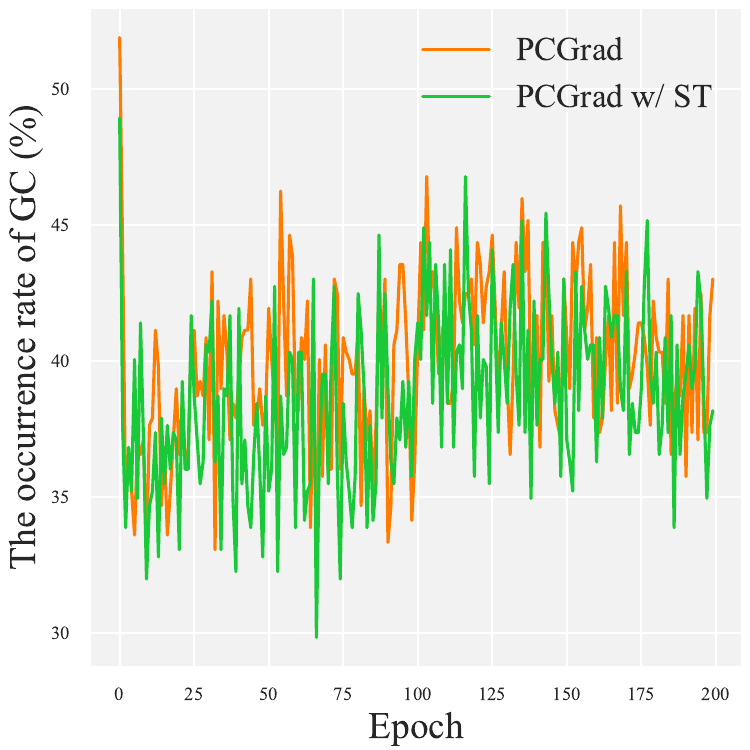}
    \label{fig:PCGrad pcgrad_gray_mtan_city}
}
\subfloat[NashMTL]{
    \includegraphics[width=0.25\linewidth]{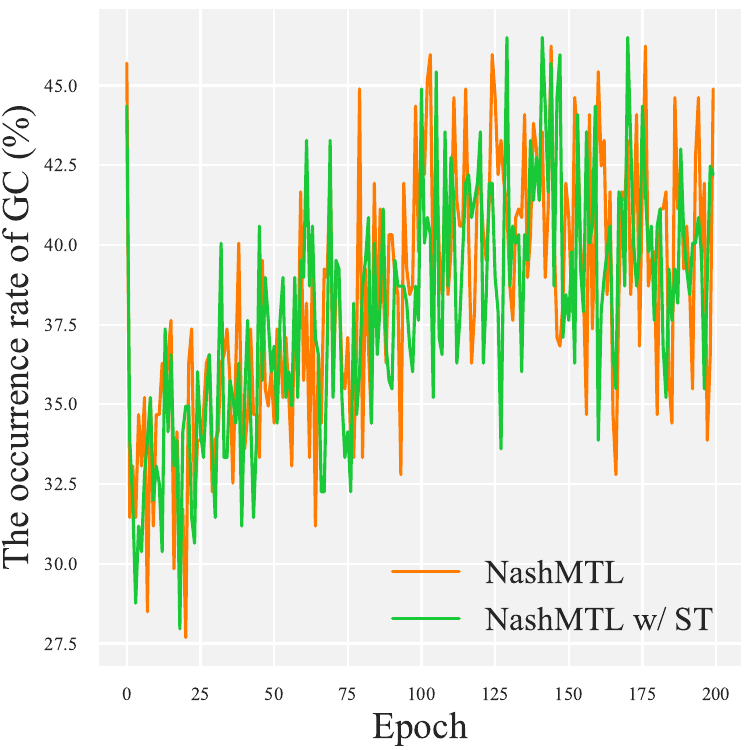}
    \label{fig:NashMTL nash_gray_mtan_city}
}\\
\subfloat[MGDA]{
    \includegraphics[width=0.25\linewidth]{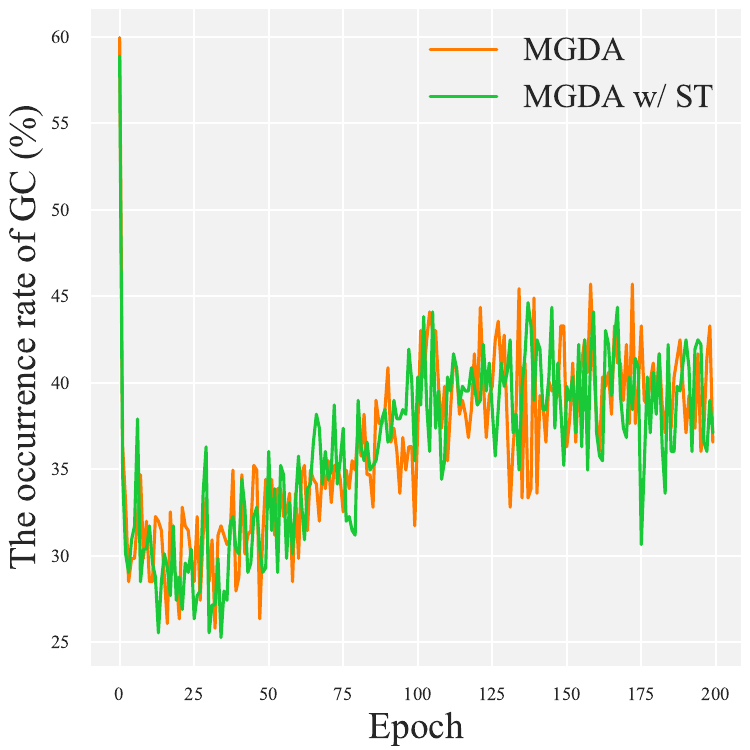}
    \label{fig:MGDA mgda_gray_mtan_city}
}
\caption{The number of occurrence gradient conflictions between tasks during tuning MTAN on CityScapes dataset.}
\label{fig:number of gc MTAN on  CityScapes} 
\end{figure*}

\begin{figure*}[t]
\centering
\includegraphics[width=1\linewidth]{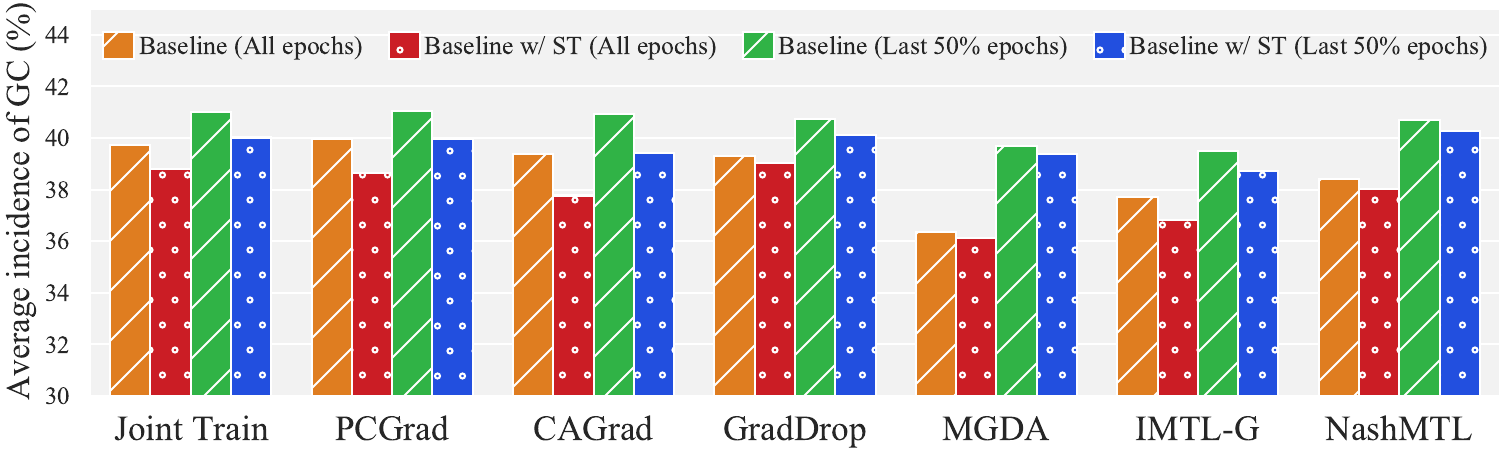}
\caption{
The average occurrence percentage of gradient conflict over epochs (all epochs/last 50\% epochs) during training on MTAN model with CityScapes datasets was evaluated using various methods, including joint training and gradient manipulation techniques.
}
\label{fig:gc of different methods CityScapes on MTAN} 
\end{figure*}

We also conduct experiments on MTAN with CityScapes dataset. MTAN is a random initialized model. As we can see in \cref{tab:gc of different methods CityScapes mtan}, even for the random initialized model, sparse training can also reduce the incidence of gradient conflict. The reduction in the incidence of gradient conflict for CityScapes is observed to be comparatively smaller than that for NYU-v2. This discrepancy can be attributed to the fact that CityScapes, which involves only two tasks, has a lower likelihood of encountering gradient conflicts between tasks compared to NYU-v2, which encompasses three tasks. The visualization of the occurrence of gradient conflict for each epoch is shown in 
\cref{fig:number of gc MTAN on  CityScapes} and the average incidence of gradient conflict across all epochs for different methods is shown in \cref{fig:gc of different methods CityScapes on MTAN}
. As for the performance of the overall tasks on CityScapes, the sparse training improves all methods including \textit{Joint Train} and all gradient manipulation methods, as shown in \cref{tab:city mtan}.

\begin{table*}[t]
    \centering
    \resizebox{0.6\textwidth}{!}{%
    \begin{tabular}{ccccccccccccccc}
    \toprule 
     \multirow{2}{*}{\textbf{Methods}}   & & \multicolumn{2}{c}{Segmentation} & \multicolumn{2}{c}{Depth}  & \multirow{2}{*}{$\mathbf{\Delta m \%} \downarrow$} \\
     \cmidrule(lr){3-4} \cmidrule(lr){5-6} 
      &  & \multirow{1}{*}{mIoU $\uparrow$} & \multirow{1}{*}{Pix Acc $\uparrow$}  & \multirow{1}{*}{Abs Err $\downarrow$} & \multirow{1}{*}{Rel Err $\downarrow$} \\
    \toprule
    \multicolumn{2}{c}{STL}& 77.61&94.15&0.0122&35.68& $-$   \\
    \midrule
    \multicolumn{2}{c}{Joint Train} &78.14&94.29&0.0174&59.21&26.87\\
    \multicolumn{2}{c}{w/ ST} &\cellcolor[HTML]{CCFFCC}78.34&\cellcolor[HTML]{CCFFCC}94.34&\cellcolor[HTML]{CCFFCC}0.0143&\cellcolor[HTML]{CCFFCC}55.00&\cellcolor[HTML]{CCFFCC}17.48 \\
    \midrule
    \multicolumn{2}{c}{PCGrad} &77.79&94.21&0.0155&51.99&19.96\\
    \multicolumn{2}{c}{w/ ST} &\cellcolor[HTML]{CCFFCC}77.79&\cellcolor[HTML]{CCFFCC}94.26&\cellcolor[HTML]{CCFFCC}0.0160&\cellcolor[HTML]{CCFFCC}51.99&\cellcolor[HTML]{CCFFCC}19.22 \\
    \midrule
    \multicolumn{2}{c}{CAGrad}    &76.82&93.70&0.0138&53.74&16.26  \\
    \multicolumn{2}{c}{w/ ST} &\cellcolor[HTML]{CCFFCC}77.20&\cellcolor[HTML]{CCFFCC}94.01&0.0150&\cellcolor[HTML]{CCFFCC}39.85&\cellcolor[HTML]{CCFFCC}8.88  \\
    \midrule
    \multicolumn{2}{c}{GradDrop}  &77.91&94.28&0.0154&55.58&20.34 \\
    \multicolumn{2}{c}{w/ ST} &\cellcolor[HTML]{CCFFCC}78.34&\cellcolor[HTML]{CCFFCC}94.38&0.0163&\cellcolor[HTML]{CCFFCC}48.95&\cellcolor[HTML]{CCFFCC}17.45 \\
    \midrule
    \multicolumn{2}{c}{MGDA} &69.91&92.17&0.0124&40.68& 6.91 \\
    \multicolumn{2}{c}{w/ ST} &68.38&91.91&0.0128&\cellcolor[HTML]{CCFFCC}33.19&\cellcolor[HTML]{CCFFCC}3.17\\
    \midrule
    \multicolumn{2}{c}{IMTL-G}     &77.55&94.10&0.0135&47.17&10.65\\
    \multicolumn{2}{c}{w/ ST} &75.75&93.98&0.0138&\cellcolor[HTML]{CCFFCC}40.16&\cellcolor[HTML]{CCFFCC}7.10\\
    \midrule
    \multicolumn{2}{c}{NashMTL}&77.51&94.22&0.0152&36.36&6.68   \\
    \multicolumn{2}{c}{w/ ST} &76.87&94.09&\cellcolor[HTML]{CCFFCC}0.0148&\cellcolor[HTML]{CCFFCC}33.30&\cellcolor[HTML]{CCFFCC}3.99  \\
    \bottomrule  
    \end{tabular}
    }
    \caption{The test performance on CityScapes dataset training on MTAN model. The green cell color indicates that sparse training improves the performance of joint training or gradient manipulation methods. The best result is highlighted in bold.}
    \label{tab:city mtan}
\end{table*}

\begin{table*}[t]
        \centering
        \resizebox{0.5\textwidth}{!}{
        \begin{tabular}{ccll}
        \toprule 
         & \multirow{2}{*}{\textbf{Methods}}& \multicolumn{2}{c}{Average incidence of GC ($\%$)}\\
         \cmidrule(lr){3-4} &                                  &All epochs& Last 50\% epochs\\
        \toprule
        &Joint Train& 39.72 & 40.99\\
        &     w/ ST &  38.79 (\textcolor{blue}{0.93})& 40.02 (\textcolor{blue}{0.97})\\
        \midrule
         & PCGrad &39.98&41.06 \\
         & w/ ST &38.66(\textcolor{blue}{1.32})&39.97(\textcolor{blue}{1.09})\\
        \midrule
         & CAGrad&39.39&40.94\\
         & w/ ST&37.77(\textcolor{blue}{1.62})&39.42(\textcolor{blue}{1.52})\\
        \midrule
         & GradDrop &39.32&40.72\\
         & w/ ST &  39.03(\textcolor{blue}{0.29})&40.12(\textcolor{blue}{0.60})\\
        \midrule
         & MGDA & 36.37&39.69\\
         & w/ ST& 36.14(\textcolor{blue}{0.23})&39.38(\textcolor{blue}{0.31})\\
        \midrule
         & IMTL-G &37.72&39.51\\
         & w/ ST & 36.83(\textcolor{blue}{0.89})&38.72(\textcolor{blue}{0.79})  \\
        \midrule
         & NashMTL &38.40&40.69 \\
         & w/ ST & 38.04(\textcolor{blue}{0.36})&40.26(\textcolor{blue}{0.43})   \\
        \bottomrule
        \end{tabular}%
        }
        \caption{Average incidence of gradient conflict between tasks over epochs for different methods. The improvement by sparse training is provided in (\textcolor{blue}{$\bullet$}). We calculate the average incidence of gradient conflict over all epochs and the last 50\% epochs during training MTAN on CityScapes.} 
        \label{tab:gc of different methods CityScapes mtan} 
\end{table*}

\subsection{FAMO}
FAMO \cite{liu2023famo} is an approximation method for gradient manipulation by using the history of loss to compute the current task weight. We also try our sparse training with FAMO on NYU-v2, CelebA, Clevr, SmallORB datasets with ViT, SAM, MTAN and Swin models. As shown in \cref{tab:nyu sam FAMO}, \cref{tab:nyu mtan FAMO}, \cref{tab:smallnorb FAMO} and \cref{tab:celeba clevr nyuv2 city FAMO}, even for the approximation method, sparse training method achieves the best results and further show the effectiveness of our sparse training methods.

\begin{table*}[t]
\centering
\resizebox{1\textwidth}{!}{
\begin{tabular}{ccccccccccccccccccccc}
\toprule 
 & \multirow{3}{*}{\textbf{Methods}}   &  \multicolumn{2}{c}{Segmentation} &   \multicolumn{2}{c}{Depth}   & \multicolumn{5}{c}{Surface Normal} &    \multirow{3}{*}{$\mathbf{\Delta m \%} \downarrow$}  \\
 \cmidrule(lr){3-4} \cmidrule(lr){5-6} \cmidrule(lr){7-11}
 &  &  \multirow{2}{*}{mIoU $\uparrow$} & \multirow{2}{*}{Pix Acc $\uparrow$}   & \multirow{2}{*}{Abs Err $\downarrow$} & \multirow{2}{*}{Rel Err $\downarrow$}  & \multicolumn{2}{c}{Angle Distance $\downarrow$}   & \multicolumn{3}{c}{Within $t^\circ$  $\uparrow$} & &  &  \\
 \cmidrule(lr){7-8} \cmidrule(lr){9-11}
 &  &  &  &  &  & Mean & Median  & 11.25 &  22.5 &  30 &    \\
\toprule
 & \multicolumn{1}{c}{FAMO} & $57.64$ & $78.59$ & $0.3574$ & $0.1463$ & $19.396$ & $12.846$ &  $45.61$ & $71.87$ & $80.59$ & $-0.5669$\\
 & \multicolumn{1}{c}{w/ ST} & \cellcolor[HTML]{CCFFCC}$57.68$ & \cellcolor[HTML]{CCFFCC}$78.79$ &\cellcolor[HTML]{CCFFCC}$0.3520$ & \cellcolor[HTML]{CCFFCC}$0.1430$& \cellcolor[HTML]{CCFFCC}$19.279$ & \cellcolor[HTML]{CCFFCC}$12.711$ &  \cellcolor[HTML]{CCFFCC}$46.12$ &\cellcolor[HTML]{CCFFCC}$72.06$ & \cellcolor[HTML]{CCFFCC}$80.72$ & \cellcolor[HTML]{CCFFCC}$-1.353$\\
\bottomrule
\end{tabular}
}
\caption{The test performance on NYU-v2 dataset training on SAM model. The green cell color indicates that sparse training improves the performance of joint training or gradient manipulation methods. The best result is highlighted in bold.}
\label{tab:nyu sam FAMO}
\end{table*}

\begin{table*}[t]
    \centering
    \resizebox{0.5\textwidth}{!}{
    \begin{tabular}{cccccccc}
    \toprule 
     &  &  \multicolumn{2}{c}{\textbf{CelebA}} & \multicolumn{3}{c}{\textbf{Clevr}}  & \multicolumn{1}{c}{\textbf{NYU-v2}}\\
     \cmidrule(lr){3-4} \cmidrule(lr){5-7} \cmidrule(lr){8-8} 
     & \multirow{1}{*}{\textbf{Methods}}& & \multirow{1}{*}{$\mathbf{\Delta m \%} \downarrow$} &\multicolumn{1}{c}{Counting}  & \multicolumn{1}{c}{Depth} & \multirow{2}{*}{$\mathbf{\Delta m \%} \downarrow$}& \multirow{2}{*}{$\mathbf{\Delta m \%} \downarrow$}\\
      \cmidrule(lr){5-5} \cmidrule(lr){6-6} 
     &  &  & (F1) & \multicolumn{1}{c}{(Top 1 $\uparrow$)}  &   \multicolumn{1}{c}{(Top 1 $\uparrow$)} & \\
    \toprule
    & \multicolumn{2}{c}{FAMO} &$2.35$& $55.83$ & $56.80$ & $3.16$&-4.10  \\
    & \multicolumn{2}{c}{w/ ST} &\cellcolor[HTML]{CCFFCC}$2.32$&\cellcolor[HTML]{CCFFCC}$\textbf{62.57}$   & $56.04$&  \cellcolor[HTML]{CCFFCC}$\textbf{-1.93}$ &\cellcolor[HTML]{CCFFCC}-4.46\\
    \bottomrule
    \end{tabular}
    }
    \caption{The test performance on CelebA, Clevr and NYU-v2 dataset. CelebA is trained on Swin Transformer and Clevr is trained on ViT. NYU-v2 is trained on MTAN. The green cell color indicates that sparse training improves the performance of joint training or gradient manipulation methods. The best result is highlighted in bold. }
     \label{tab:celeba clevr nyuv2 city FAMO}
\end{table*}

\begin{table*}[h]
    \centering
    \resizebox{\textwidth}{!}{%
    \begin{tabular}{ccccccccccccccc}
    \toprule 
     \multirow{3}{*}{\textbf{Methods}}   & & \multicolumn{2}{c}{Segmentation} & \multicolumn{2}{c}{Depth}  & \multicolumn{5}{c}{Surface Normal} &  \multirow{3}{*}{$\mathbf{\Delta m \%} \downarrow$} \\
     \cmidrule(lr){3-4} \cmidrule(lr){5-6} \cmidrule(lr){7-11}
      &  & \multirow{2}{*}{mIoU $\uparrow$} & \multirow{2}{*}{Pix Acc $\uparrow$}  & \multirow{2}{*}{Abs Err $\downarrow$} & \multirow{2}{*}{Rel Err $\downarrow$}   & \multicolumn{2}{c}{Angle Distance $\downarrow$} & \multicolumn{3}{c}{Within $t^\circ$  $\uparrow$} \\
     \cmidrule(lr){7-8} \cmidrule(lr){9-11}
     &  &  &  &  &  & Mean & Median  & 11.25 & 22.5  & 30 \\
    \toprule
    \multicolumn{2}{c}{FAMO}      & 38.88 & 64.90 & 0.5474 & 0.2194 & 25.06 & 19.57 & 29.21 & 56.61 & 68.98   & -4.10\\
    \multicolumn{2}{c}{w/ ST} &37.85 &\cellcolor[HTML]{CCFFCC}65.27 &0.5543 &0.2215 &25.09 &\cellcolor[HTML]{CCFFCC}19.15 &\cellcolor[HTML]{CCFFCC}30.03 &\cellcolor[HTML]{CCFFCC}57.49 &\cellcolor[HTML]{CCFFCC}69.52 &\cellcolor[HTML]{CCFFCC}-4.46 \\
    \bottomrule  
    \end{tabular}
    }
    \caption{The test performance on NYU-v2 dataset training on MTAN model. The green cell color indicates that sparse training improves the performance of joint training or gradient manipulation methods. The best result is highlighted in bold.}
    \label{tab:nyu mtan FAMO}
\end{table*}

\begin{table*}[h]
\centering
\begin{tabular}{ccccccccccccc}
\toprule 
 & \multirow{2}{*}{\textbf{Methods}}& &\multicolumn{2}{c}{Object Azimuth} &  & \multicolumn{2}{c}{Camera Elevation} & \multirow{2}{*}{$\mathbf{\Delta m \%} \downarrow$}\\
 \cmidrule(lr){4-5} \cmidrule(lr){7-8} \cmidrule(lr){10-12}
 &  &  & Top 1 $\uparrow$ &Top 5 $\uparrow$ &  &  Top 1 $\uparrow$ & Top 5 $\uparrow$ \\
\toprule
& \multicolumn{2}{c}{FAMO} & $24.68$& $63.69$& & $34.35$&$92.13$ & $10.71$  \\
& \multicolumn{2}{c}{w/ ST} &\cellcolor[HTML]{CCFFCC}$26.38$ & \cellcolor[HTML]{CCFFCC}$66.54$& & $32.05$& $91.02$&  \cellcolor[HTML]{CCFFCC}$10.27$\\
\bottomrule
\end{tabular}%
\caption{The test performance on SmallNORB dataset trained on ViT. The green cell color indicates that sparse training improves the performance of joint training or gradient manipulation methods. The best result is highlighted in bold. }
\label{tab:smallnorb FAMO}
\end{table*}

\end{document}